\documentclass[11pt]{article}
\usepackage[UKenglish]{babel}
\usepackage[a4paper, margin=1in]{geometry}

\usepackage{hyperref}
\usepackage{graphicx}
\usepackage{caption}
\usepackage{subcaption}
\usepackage{amsmath}
\usepackage{amssymb}
\usepackage{mathtools}
\usepackage{siunitx}
\usepackage{multirow}
\usepackage{xcolor}
\usepackage{ctable}
\usepackage{bm}

\DeclareMathOperator*{\argmin}{arg\,min}
\DeclareMathOperator*{\argmax}{arg\,max}

\graphicspath{{figures/}}

\title{Digital Fingerprinting of Microstructures}
\author{M.D. White$^{a,}$\footnote{Corresponding author \newline \textit{Email address:} michael.white-3@postgrad.manchester.ac.uk (M.D. White)} \ , A. Tarakanov$^{b}$, C.P. Race$^{a}$, P.J. Withers$^{a}$, K.J.H. Law$^{b}$ \\ \\
\small $^{a}$\textit{Henry Royce Institute, Department of Materials, University of Manchester, M13 9PL, UK} \\
\small $^{b}$\textit{Department of Mathematics, University of Manchester, M13 9PL, UK}}
\date{\small \today}

\begin{document}

\maketitle

\noindent\rule{\textwidth}{.5pt}
\begin{abstract}
    Finding efficient means of quantitatively describing material microstructure is a critical step towards harnessing data-centric machine learning approaches to understanding and predicting processing-microstructure-property relationships.
    Current quantitative descriptors of microstructure tend to consider only specific, narrow features such as grain size or phase fractions, but these metrics discard vast amounts of information.
    Since the gain in traction of machine learning and computer vision, more abstract methods for describing image data in a concise and quantitative manner have become available, but have yet to be fully exploited within materials science.
    The main aim of this paper is to investigate some of these methods as tools for constructing compressed numerical descriptions of microstructural image data, which are here referred to as ``microstructural fingerprints''.
    A statistical framework is developed to combine some of these methods which includes some classical computer vision methods as special cases.

    The effectiveness of fingerprints, in this case, is assessed via a series of classification tasks, which take the fingerprints as input along with some label to describe the microstructure, and aim to predict the class label.
    The classification tasks are a simple way to assess the information content of differently constructed fingerprints, using readily available data.
    However, the potential applications for such fingerprints, in theory, extend far beyond this, provided a suitably labelled training dataset.
    For example, fingerprints constructed from image data labelled with some processing parameters or mechanical properties could be fed into regression-based tasks to predict properties form microstructure or predict the processing parameters required for a target microstructure.
    The ultimate purpose is to rapidly fingerprint sample images in the context of various high-throughput design/make/test scenarios.
    Such fingerprints would enable, for example, quantification of the disparity between microstructures for quality control, classifying microstructures, predicting materials properties from image data and identifying potential processing routes to engineer new materials with specific properties.

    Here, the approach is applied to two distinct datasets to illustrate various aspects of the fingerprints and some recommendations are made based on the findings.
    In particular, methods that leverage transfer learning with convolutional neural networks (CNNs), pretrained on the ImageNet dataset, are generally shown to outperform other methods.
    Additionally, dimensionality reduction of these CNN-based fingerprints is shown to have negligible impact on classification accuracy for the supervised learning approaches considered.
    In situations where there is a large dataset with only a handful of images labelled, graph-based label propagation to unlabelled data is shown to be favourable over discarding unlabelled data and performing supervised learning.
    In particular, label propagation by Poisson learning is shown to be highly effective at low label rates.
    The supplementary code is available on GitHub~\cite{mftools}.
\end{abstract}
\noindent\rule{\textwidth}{.5pt}

\newpage
\section{Introduction}\label{sec:intro}

Materials informatics, namely the use of data science and informatics approaches to improve processing and performance of materials, has advanced significantly in recent years, see e.g.~\cite{kalidindi2015hierarchical,agrawal2016perspective,holm4}.
More recently, microstructural informatics, the process of collecting, organising, sharing and simulating microstructural data and relating microstructure to processing parameters and properties has garnered interest, particularly since the the uptake in machine learning (ML).

A material's microstructure can be viewed as a platform to build bridges between processing and properties: processing determines the microstructure which, in turn, determines mechanical properties.
Correlations could be sought directly between processing and properties, however, it has been shown that leveraging microstructural information can accelerate and enhance model performance for material design~\cite{molkeri2022}.
Including details of the microstructure in the inferential chain also mirrors the way that metallurgy is traditionally practised and so will improve interpretability and hence acceptance of new methods.
Whilst some microstructural information, in the form of various metrics, is often included in such models, there is little discussion around the optimal methods for quantitatively describing microstructure more generally and how this affects model performance.
Here, we refer to a general quantitative representation of microstructure as a ``microstructural fingerprint''.
There are already several options for fingerprint construction within the literature, although these representations are often viewed as a means of improving the accuracy of a specific ML task, rather than seeking a truly representative compressed form of microstructural image data (which could have universal applications).
The question is then whether such representations can describe the ``essence'' of the microstructure itself, or just a particular image.
A traditional approach would be to make measurements of metrics such as grain size/distribution, lamellar spacing etc.
and concatenate these metrics into a vector to form the fingerprint~\cite{schmitz2016}.
This is shown to be useful and can predict properties such as tensile strength~\cite{xie2021}.
However, this approach is rather limited by the number of metrics at our disposal and, also, by the fact that different metrics apply in each case, depending on the microstructure.
Two-point correlation functions acquired from binary micrographs, reduced via principal component analysis (PCA),
have been shown to provide a useful representation of microstructure and can also be applied to predict elastic properties~\cite{ford2021, yucel2020}.
Prediction of elastic properties in the form of full stress-strain curves has also been shown to be possible, with the use of synthetic pole figures constructed from electron backscatter diffraction (EBSD) data as input~\cite{yamanaka2020}.
Generative algorithms, such as generative adversarial networks (GANs) have been shown to enable construction of synthetic datasets~\cite{ma2020image,lambardo2020}, which is very useful in its own right, but the compressed representations defined by the latent space which initialise the GAN may also be considered as fingerprints and are yet to be fully exploited.

Finding efficient means of fingerprinting microstructure is critical if we are to capitalise on data-centric learning approaches~\cite{withers2019}.
Within the last decade, there has been a drive towards the identification and analysis of keypoint features in microstructural images~\cite{holm1,holm2}
and subsequent supervised learning (SL) tasks~\cite{holmdnn,azimi2018advanced}, which operate with a set of training data labelled with relevant properties.
There are a growing number of works on unsupervised learning (UL), which can operate with the same features as SL, but without any labels~\cite{holm2,holm3,holm4,mueller2021machine,ma2020image}.
A natural addition to these are the semi-supervised learning (SSL) methods, which operate on a set of $N$ inputs with a set of $pN$ labels, where $p \in (0,1)$.
SSL can be regarded as spanning the range from UL ($p=0$) to SL ($p=1$).
A few works on SSL have appeared in recent years~\cite{ssl1,ssl2,ssl3,ssl4}.
The present work adds to the previous work on microstructural fingerprinting by (i) presenting a unified framework for fingerprinting, which includes classical feature extraction methods from computer vision known as visual bag of words (VBOW)~\cite{filliat2007visual,csurka2004visual,lazebnik2005sparse,nowak2006sampling,holm1} and
vector of locally aggregated descriptors (VLAD)~\cite{jegou2010aggregating,delhumeau2013revisiting,holm2},
and more recent transfer learning with pretrained
convolutional neural networks (CNNs)~\cite{cimpoi2015deep,jia2014caffe,parkhi2014compact,holm2,holm3,holm4} and
(ii) exploring the potential utility of some graph-based SSL methods~\cite{bertozzi2012diffuse,calder2020poisson} in this context, which can be applied to large datasets where only a small percentage of the data is labelled.
This removes the requirement for large-scale labelling procedures which can be time-consuming and cumbersome.

VBOW and VLAD are based on summary statistics of what are referred to here as base features, which can be keypoint-based features, such as scale-invariant feature transform (SIFT)~\cite{lowe1999object,lowe2004distinctive} and speeded-up robust features (SURF)~\cite{bay2008speeded}, as well as more recently adopted transfer learning approaches to feature extraction based on pretrained CNNs~\cite{gong2014multi,cimpoi2015deep,holm2}.
There are by now numerous choices of CNN to use for transfer learning, and one example which has shown some success in this application, developed by the Visual Geometry Group at Oxford, is known as VGG~\cite{vgg,holm2}.

The graph-based classification methods considered aim to capture the intrinsic geometry of the data manifold and construct a low-dimensional representation, if one exists.
As such, they can be used for UL~\cite{shi2000normalized,von2007tutorial} and, in the case that a goal and some labelled data are identified a priori, SSL~\cite{bertozzi2012diffuse,calder2020poisson}.
These methods are mature by now, and there is a growing body of mathematics literature relating to convergence of
the graph Laplacian to the Laplace Beltrami operator on the feature space
and consequences for the various related algorithms~\cite{shi2000normalized,von2007tutorial}.
It is shown here that, for a priori known classification tasks, these SSL methods are able to significantly improve upon the accuracy of UL methods with very few labels (as few as $5\%$) and even approach the accuracy of SL methods.
The methods therefore represent a useful contribution in the context of microstructure characterisation (as well as other imaging domains), where the amount of data is vast, but the process of labelling is costly.

In this study, we present a comparison between local keypoint and CNN methods for base feature extraction and fingerprint construction from said features.
We begin by outlining the general approach for fingerprint construction from base features in Section~\ref{sec:fp_generic}.
Section~\ref{sec:base} then specifies several candidates for base feature extraction.
A framework for performing statistical analysis of base features, the final stage in fingerprint construction, is proposed in Section~\ref{sec:finger}.

Machine learning classification methods (namely SL, SSL and UL) are described in Section~\ref{sec:learning} and accuracy assessment of such methods is discussed in Section~\ref{sec:quality}, as a tool for measuring fingerprint quality (i.e. the amount of meaningful information contained in the fingerprints).
Section~\ref{sec:data} provides details of the data that was used for testing and outlines the classification task that will form the basis of our fingerprint quality assessment.
Results are then presented in Section~\ref{sec:numerics} (supplementary code can be found at~\cite{mftools}) and discussed in Section~\ref{sec:discussion}.
A concise summary of the conclusions is provided in Section~\ref{sec:conclusions}.

\section{General Approach to Fingerprinting}
\label{sec:fp_generic}

A fingerprint of an image can be considered as a compressed numerical representation of image data.
This can be considered as a feature mapping from the image to a vector, where the dimensionality of the vector is relatively small compared to the number of pixels in the image.
This can be defined more rigorously as
\begin{gather*}
    f: \mathbb{R}^{m_1 \times m_2} \rightarrow \mathbb{R}^n, \\
    x \mapsto f(x),
\end{gather*}
where $x$ is an $m_1 \times m_2$ image and $f(x)$ is the fingerprint, an $n$-dimensional vector, such that $n \ll m_1 m_2$.
Note that $m_1$ and $m_2$ can depend on $x$, i.e.\ the number of pixels can vary from image to image.
However, this is excluded to simplify the notation.
A microstructural fingerprint refers to an image fingerprint computed from a micrograph.
Here, we consider images acquired via optical microscopy and scanning electron microscopy (SEM), although all methods discussed can also apply to 3-channel colour images and, thus, there is potential to extend the application to include electron backscatter diffraction (EBSD) data, provided with a suitable dataset.

There already exists a variety of options for construction of the fingerprint, $f(x)$.
The general approach considered here is centred around constructing base features from the image and clustering said features, before applying summary statistics on the clustered features.
Base features are constructed via a map
$S: \mathbb{R}^{m_1\times m_2} \rightarrow \mathbb{R}^{J(x) \times d}$,
which delivers a set of $J(x)$ base features, $\{S_1(x), \dots, S_{J(x)}(x)\}$, for each image, where each feature is represented as a $d$-dimensional vector.
Suppose there are $N$ images.
A population, $\mathcal{P}$, of all features from all images can then be constructed, such that
\[
    \mathcal{P} = \left\{ \{S_j(x_i)\}_{j=1}^{J(x_i)} \right\}_{i=1}^N = \{S_p\}_{p=1}^{P},
\]
where $P$ denotes the total number of features in $\mathcal{P}$.

A clustering function is then learned from $\mathcal{P}$,
which assigns a label, $t_p \in \{1, \dots, K\}$, to each base feature, $S_p \in \mathcal{P}$,
resulting in ``similar'' features being assigned the same cluster label.
$k$-means clustering~\cite{sammut_encyclopedia_2010} was utilised at this step for all results presented in this paper.
$k$ here is arbitrary and refers to the number of clusters to fit to the population of features.
Cluster centres, $\mu_k$, for $k = 1, \dots, K$, are learned to minimise distance between cluster centres and data points.
The $\mu_k$ are initially randomised and subsequently updated according to
\begin{equation} \label{eq:kmeans1}
\mu_k = \frac{1}{|\mathcal{S}_k|} \sum_{S_p \in \mathcal{S}_k} S_p,
\end{equation}
where
\[
    \mathcal{S}_k \coloneqq \{S_p \in \mathcal{P}: \ t_p=k \}
\]
and cluster labels are assigned according to
\[
    t_p = \argmin_{k\in\{1, \dots ,K\}} | S_p - \mu_k |^2.
\]
This process is repeated in a loop in which each feature is assigned a cluster label according to the current nearest cluster centre.
Each cluster centre is then updated via~\eqref{eq:kmeans1}, which effectively results in the cluster centre moving towards the centre of mass of those features currently belonging to that cluster.

Spectral clustering~\cite{shi2000normalized} was considered, but was too computationally expensive to validate.
The Nystr\"{o}m extension~\cite{pourkamali-anaraki_scalable_2020} offers an approximation to spectral clustering, however, the results were very poor during testing and so this was omitted.
We will revisit spectral clustering in Section~\ref{ssec:ssl}, as it is a viable option for clustering fingerprints (rather than clustering base features) for unsupervised learning classification,
provided there is a relatively low number of classes in the dataset and/or fingerprints are relatively low-dimensional (this will depend on computation power available).

Finally, various summary statistics of $\{S_j(x)\}_{j=1}^{J(x)}$
are constructed for each sample image, $x$.
Summary statistics, of order $l$, are denoted $H_l: \mathbb{R}^{m_1 \times m_2} \rightarrow \mathbb{R}^{n_l}$ for $l=0,1,2,\dots$, where $n_l$ depends on the order of the fingerprint and the number of clusters.
One or more summary statistics are ultimately combined to form the fingerprint $f(x) \in \mathbb{R}^n$.

\section{Base Features}
\label{sec:base}

There are several options for constructing base features.
Here, we consider keypoint-based features and transfer learning features from pretrained convolutional neural networks (CNNs).
In either case, we can consider each feature as a local descriptor of a given patch from the image.
Although, in some cases (for CNN-based fingerprints), the patch may be defined as the entire image itself.

\subsection{Patch-Based Maps}
\label{ssec:patches}

Patch-based methods first identify a set of patches
$z_i \in \mathbb{R}^{m\times m}$, for $i = 1, \dots, J(x)$,
from a given image, $x$.
We then define $S_i(x) = s(z_i)$, where
$s : \mathbb{R}^{m\times m} \rightarrow \mathbb{R}^d$,
i.e. each feature can be considered as a vector describing a local neighbourhood centred at each keypoint (patch centre).

The set $\Omega(x) \coloneqq \{z_1, \dots, z_{J(x)}\}$ can be constructed
according to a particular span and stride,
e.g. if one has square images ($m_1=m_2$) and $m_1$ is divisible by $m'$, then letting $m=m_1/m'$ and selecting vertical and horizontal
stride of $m$ produces a partition of $x$ into $(m')^2$ patches, $z \in \mathbb{R}^{m\times m}$.
If one allows $\Omega(x)$ to be random, then it could consist of patches around randomly selected points,
which can be viewed as a version of bootstrapping for data augmentation.

\subsubsection{Keypoint Features}
\label{ssec:sift}

The set $\Omega(x)$ can also be designed using keypoint-based methodology,
which has proven very valuable in the field of computer vision,
particularly for object recognition.
Such methods identify a set of $J(x)$ keypoints from an image, $x$,
together with orientation and scale information.
This is used together with interpolation to define the patches.
One could, in principle, ignore the scale or orientation information, or both.
There are several candidates for keypoint feature extraction.
Here, we consider the scale-invariant feature transform (SIFT)~\cite{lowe2004distinctive} and a lower-dimensional alternative, speeded-up robust features (SURF)~\cite{bay2008speeded}.

\begin{figure}[!ht]
    \centering
    \includegraphics[width=\textwidth]{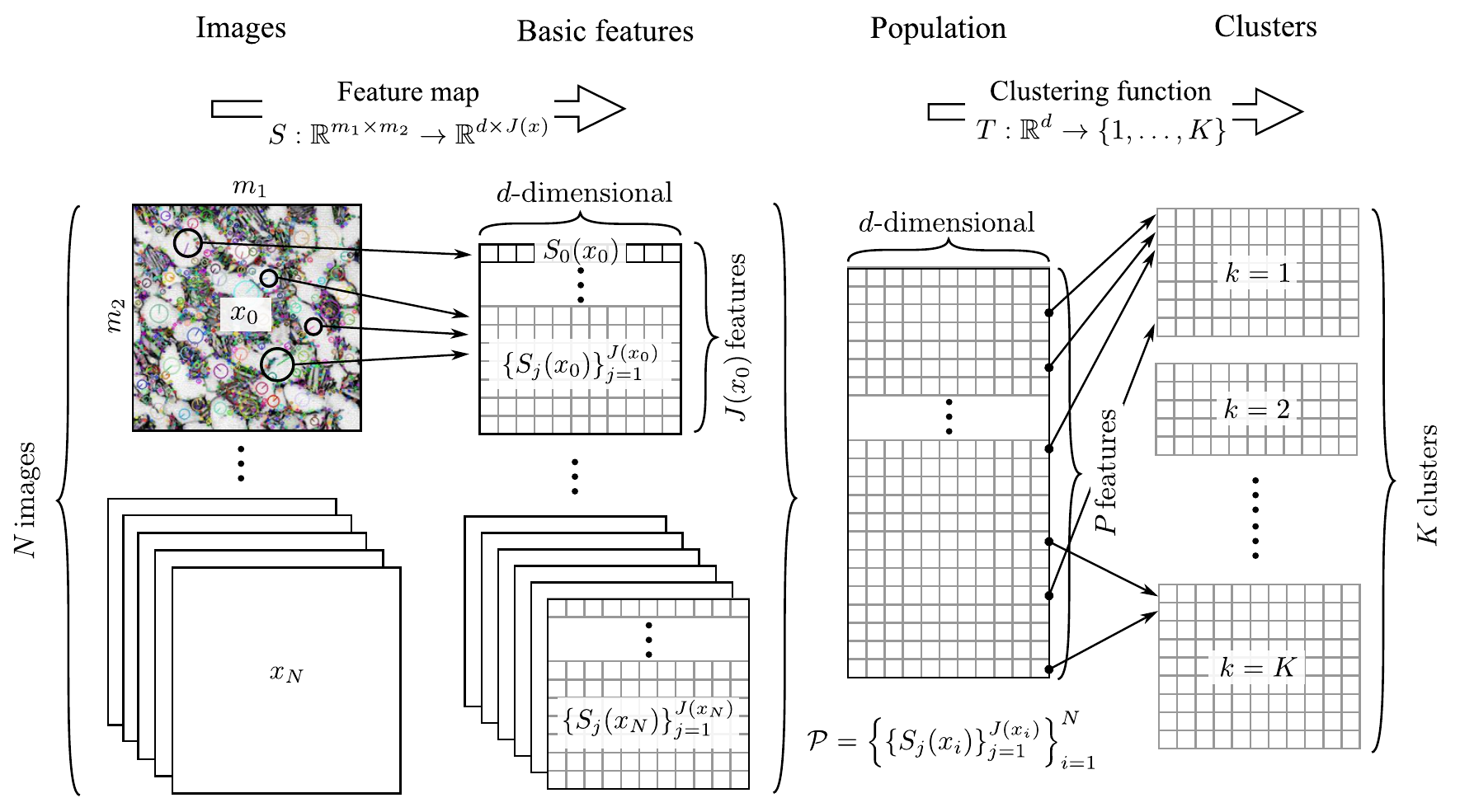}
    \caption{Visual representation of feature population construction and clustering from SIFT keypoint features, where each feature is represented as a ring, centred at a keypoint, with size and orientation corresponding to the scale at which the feature was detected and the dominant orientation of the feature, respectively.}
    \label{fig:sift}
\end{figure}

The SIFT algorithm detects keypoints via Difference of Gaussian (DoG).
Gaussian blur is applied to the input image several times, each to varying degrees.
The blurred images are then stacked and the DoG is computed via Euclidean distance between adjacent Gaussians.
Local extrema are identified and their locations serve as the keypoints.
A $16 \times 16$ neighbourhood is then constructed, centred at each keypoint, and divided into 16 $4 \times 4$ subregions.
The scale of this neighbourhood depends on the Gaussian scale at which the keypoint was detected.
An 8-bin histogram of orientation gradients (HOG) is then computed for each of the 16 subregions and concatenated to form a 128-dimensional feature vector for each keypoint.
Inclusion of this orientation information ensures the features are rotation invariant.
For further discussion of the finer properties of these methods,
the reader is referred to~\cite{lowe2004distinctive}.
The top left of Figure~\ref{fig:sift} shows an example of a micrograph annotated with SIFT descriptors that are represented by coloured rings.
The size of these rings correspond to the Gaussian scale at which the keypoint was detected and the dominant orientation is represented by a straight line emanating from the centre of the ring.
The remainder of the figure provides a visual representation of feature population, $\mathcal{P}$, construction and subsequent clustering.

SURF descriptors are similar to SIFT, in the sense that they also contain information regarding gradient distribution.
However, they are much faster to compute and benefit (as far as compression is concerned) from being only 64-dimensional vectors, compared to SIFT which produces features of twice that size.
To achieve this, the DoG is first approximated via convolution with box filters, where the size of the filter corresponds to the amount of Gaussian blur in the DoG.
Keypoints are then identified as the location of local extrema across scales, as with SIFT.

Keypoints are described locally via the Haar wavelet transform~\cite{van_fleet_haar_2011}.
A local neighbourhood is taken around each keypoint and split into 16 subregions, which form a $4 \times 4$ grid centred at the keypoint.
The Haar wavelet response is then recorded for each subregion as a 4-dimensional vector, $[dx, |dx|, dy, |dy|]$, where $dx$, $dy$ is the Haar wavelet response in the horizontal and vertical direction, respectively, and $|dx|$, $|dy|$ are the absolute values of the responses.
Haar wavelet responses are computed for all subregions and concatenated into a single vector.
A 4-dimensional vector for each of the 16 subregions results in the 64-dimensional descriptor.
See~\cite{bay2008speeded} for more details.

\subsection{CNN Features}
\label{ssec:cnn_feat}

Convolutional neural networks (CNNs) have proven very successful in computer vision
tasks, most notably perhaps classification of the ImageNet dataset~\cite{ILSVRC15}.
In this study, AlexNet~\cite{alexnet} and VGG19~\cite{vgg}
(pretrained on the ImageNet dataset~\cite{deng2009imagenet}) are utilised.
An extensive collection of pretrained CNN architectures are available.
These two were selected somewhat arbitrarily, the main aim being to compare a relatively low-dimensional network (AlexNet) with a deeper network that results in higher-dimensional output features (VGG).
Using the recommended input image size ($224 \times 224$) yields a 9126-dimensional output from AlexNet and a 25088-dimensional output from VGG19.

It is important to note here that the input size of a pretrained CNN is fixed by the first fully connected layer~\cite{chatfield2014a,gong2014multi,cimpoi2015deep,holm2}.
The first fully connected layer delineates the beginning of the classification portion of the CNN.
As we are using pretrained networks, the architecture of this classification portion is specific to classification of the ImageNet dataset and can be discarded to leave just the feature extraction section of the CNN.
Therefore, the output of any convolution layer can be computed for an input image of arbitrary size and recommended input sizes can be safely ignored.
In fact, forcing the recommended input size can cause problems.
There are two options for image transformation - cropping and deforming.
Cropping the image may result in a non-representative volume element, i.e. too much data is discarded.
Stretching will deform the grain morphology and at different rates across the image set dependent on variations in scale.
Again, this can mean that the preprocessed image is not representative of the microstructure itself.
However, preprocessing inputs by converting to RGB and normalising relative to the ImageNet data mean and variance across each RGB channel is crucial, as the weights associated with each CNN layer are trained to extract features from image data distributed in such a way.

Inputting raw images (not transformed in any way) yields an output from the final CNN layer with its size dependent on the size of the input image.
This output will be a tensor of size $d_1(x) \times d_2(x) \times d$, where
$d_1(x)$, $d_2(x)$ depend on both the size of the input image and CNN architecture and $d$ depends only on the CNN architecture.
We can then define the feature matrix, $S$, by these $J(x)$ $d$-dimensional feature vectors, where $J(x)=d_1(x)d_2(x)$, from the final convolution layer
and either max pool, resulting in a single $d$-dimensional output for an image, flatten in to a single $d_1(x) d_2(x) d$-dimensional feature vector,
or utilise the $H_l$ framework of summary statistics described in Section~\ref{sec:finger}, treating each row in the $J(x) \times d$ output as essentially a keypoint feature discussed in Section~\ref{ssec:sift}.

\subsection{Reduced Feature Population}
\label{ssec:red_feat}

Depending on which method is utilised for base feature extraction, the number of features extracted for each image may be relatively large, i.e. $J(x) \gg d$.
The larger $J(x)$ is, the more information we have at our disposal to describe each image.
Intuition says that maximising feature information for each image should result in the highest classification accuracy, at the cost of longer run times.
However, it is possible to reduce each individual image's feature set from a $J(x) \times d$ array to a $d \times d$ array (i.e. reduce a set of $J(x)$ features to a set of $d$ features), before concatenating to form the population of features, $\mathcal{P} \in \mathbb{R}^{dN \times d}$, where $N$ denotes the number of images in the dataset.
One approach to this is to transpose the feature set before applying principal component analysis (PCA) and transposing back again.
This results in a reduction in the number of features, rather than in the dimensionality of each feature, which is the usual aim with PCA.
Population compression can be large in some cases, and retains important information from across the entire feature array for each image, whilst speeding up the clustering process.

\section{Fingerprint Framework}
\label{sec:finger}

We propose a general framework, derived from statistical moments, for constructing fingerprints from clustered base features.
We denote the fingerprint by $H_l$, where $l$ is the order of the fingerprint.
Cases $l = 0, 1, 2$ are defined below and higher order moments follow analogously, but are not tested here.

\subsection{$H_0$}
\label{ssec:h0}
\begin{figure}[!ht]
    \centering
    \includegraphics[width=\textwidth]{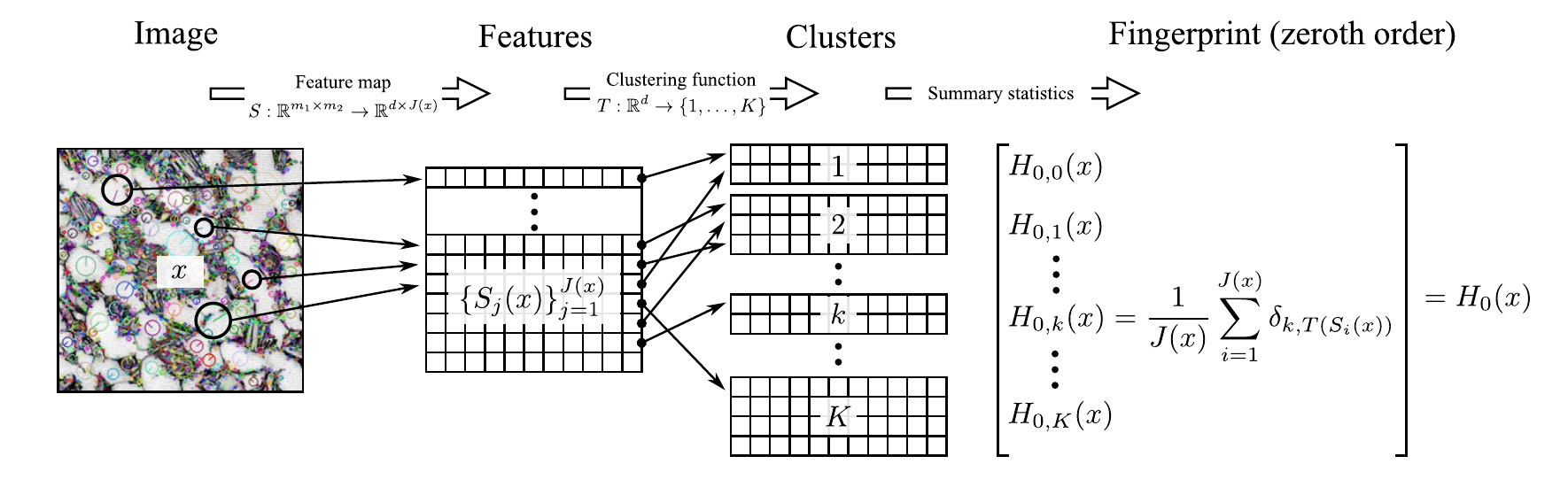}
    \caption{Schematic of $H_0$ fingerprint construction from SIFT features.}
    \label{fig:h0_schematic}
\end{figure}

The zeroth order fingerprint, $H_0$, is constructed as a histogram of feature occurrences, where each bin represents a group of ``similar'' features (defined by $k$-means clustering).
This results in a $K$-dimensional vector, where $K$ is the number of clusters and each entry in the vector denotes the number of features assigned to that cluster centre (up to normalisation).
This can be defined more rigorously as
\[
    H_{0,k}(x) = \frac{1}{J(x)}\sum_{i=1}^{J(x)} \delta_{k,T(S_i(x))},
\]
where
$\delta_{k,T(S_i(x))}$ is the Kronecker delta function and $k = 1, \dots, K$.
The full fingerprint is then given by
\[
    H_0(x) = [H_{0, 1}(x), \dots, H_{0, K}(x)]^{\top} \in \mathbb{R}^{K}_+.
\]
This is referred to as the visual bag of words (VBOW) method~\cite{filliat2007visual}
and has been considered in~\cite{holm1,holm2} with SIFT features.
Figure~\ref{fig:h0_schematic} shows a schematic to help visualise fingerprint construction in this manner.

\subsection{$H_1$}
\label{ssec:h1}
$H_1$ is a first order moment and considers mean features across each cluster centre.
This produces higher-dimensional fingerprints than $H_0$ ($K \times d$ rather than $K$, where $d$ is the length of each feature vector being clustered).
This is a result of retaining a full-length feature vector (the mean feature vector) for each cluster label, rather than just the number of feature assignments to each cluster label.
For $k=1,\dots, K$, define the $d$-dimensional mean feature vector associated to cluster $k$ as
\[
    H_{1,k}(x) = \frac{1}{J_k(x)} \sum_{i=1}^{J_k(x)} S^\top_{j_i}(x),
\]
such that $T(S_{j_i}(x)) = k$ for $i=1,\dots, J_k(x)$.
The full $K \times d$ fingerprint is then given by
\[
    H_1(x) = [H_{1,1}^\top(x), \dots, H_{1,K}^\top(x)]^\top \in \mathbb{R}^{K \times d}.
\]
This is a matrix where the $k^\text{th}$ row corresponds
to the sample mean of the $k^\text{th}$ cluster associated to the input image, $x$.
$H_1$ therefore does not contain information about the prevalence of features associated with the different clusters, but it can be combined with $H_0$ to form the fingerprint $F(x) = [H_0(x), H_1(x)] \in \mathbb{R}^{K \times (d+1)}$, which can be written as $f(x) = \text{vec} ( F(x)) \in \mathbb{R}^{n}$, where $n = K (d+1)$.

$H_1$ contains information about the character of the average feature at each cluster centre for a given image.
Depending on which feature extraction method is employed, some fingerprints may contain large numerical values which cause these fingerprints to dominate over others when training classifiers.
To help alleviate this, we can centre $H_{1,k}(x)$ by subtracting the $k^\text{th}$ cluster centre, $\mu_k^\top$, before $L_2$-normalising, i.e.
\[
    H_{1, k}^{\text{v}}(x) = \frac{H_{1, k} - \mu_k^\top}{|H_{1, k}(x) - \mu_k^\top|}.
\]
This is then exactly the vector of locally aggregated descriptors (VLAD)~\cite{jegou2010aggregating,delhumeau2013revisiting}.
Such fingerprints have been considered before in the context of
microstructure classification in~\cite{holm2},
and are shown to perform better than $H_0$.
However, it is worth noting that $n=Kd$
in comparison to $n=K$ above.
Therefore, some balance should be struck between
complexity and accuracy.
VLAD fingerprints will be denoted $H_{1}^\text{v}$.

\subsection{$H_2$}
\label{ssec:h2}
The second moment, $H_2$, is constructed similarly and is effectively encoding variance of features across each cluster.
These are higher-dimensional again (now $K \times d \times d$ for full $H_2$),
due to the fact that we are now considering distances between features belonging to each cluster centre.
This results in the $1 \times d \times d$ variance matrix associated to cluster $k$, given by
\[
    H_{2,k}(x) = \frac{1}{J_k(x)} \sum_{i=1}^{J_k(x)}
    (S_{j_i}(x) - H_{1,k}^\top(x)) \otimes (S_{j_i}(x) - H_{1,k}^\top(x)),
\]
which then leads to the full $K \times d \times d$ fingerprint of level $l = 2$ given by
\[
    H_2(x) = [H_{2,1}(x), \dots, H_{2,K}(x)] \in \mathbb{R}^{K \times d \times d}.
\]
In general, the full fingerprint $H_2(x) \in \mathbb{R}^{K \times d \times d}$ is a third order tensor.
It is possible to take the diagonal to reduce this, in which case $H_2(x)$ can be
summarised as a $K \times d$ matrix, like $H_1(x)$.
Nonetheless the dimensionality quickly grows, along with the finer information in higher moments.
We will also consider a VLAD-like second order moment,
where $(S_{j_i}(x) - H_{1,k}^\top(x))$ above is replaced with
$(S_{j_i}(x) - \mu_k)/|H_{1,k}(x) - \mu_k^\top|$, and the resulting fingerprint is denoted
$H_{2}^\text{v}(x)$.

\subsection{Multi-Scale $H_l$ Fingerprints}
\label{ssec:ms}

For a fingerprint of order $l$ with $K$ cluster centres applied to an image $x$,
the single-scale $H_l$ framework provides fingerprints described by the map
\[x \mapsto [H_{l, 1}(x), \dots, H_{l, K}(x)]^\top.\]
Multi-scale fingerprints can then be achieved through concatenation of
fingerprints with different numbers of cluster centres, i.e.
\[x \mapsto [H_{l, 1}(x), \dots, H_{l, K_1}(x), H_{l, 1}(x), \dots, H_{l, K_2}(x)]^\top,\]
where $K_1 \neq K_2$ denote the number of cluster centres utilised for two individual cluster models.
To extend this approach, an arbitrary number of cluster models can be included by considering $K_i$ for any $i > 2$.
This sort of multi-scale fingerprint might prove useful in situations where several different descriptors are required to capture different classes of features, or where a correlative approach to microstructure characterisation provides multiple images with the same field of view via different techniques.

\subsection{CNN Fingerprints}
\label{ssec:cnn_fing}

As discussed in Section~\ref{ssec:cnn_feat}, pretrained CNNs can be utilised for image-based feature extraction, by considering the final convolution layer of the feature extraction portion of the network.
The output from this layer is a tensor of size $d_1(x) \times d_2(x) \times d$, where $d_1(x), d_2(x)$ depend on both CNN architecture and input image size and $d$ depends just on CNN architecture.
We consider three methods for fingerprint construction from this tensor.
\begin{enumerate}
    \item Flatten the feature tensor into a single vector, resulting in a $d_1(x) d_2(x) d$-dimensional fingerprint.
    \item Max pool each $d_1(x) \times d_2(x)$ patch into a $d$-dimensional fingerprint.
    \item Reshape into a $J(x) \times d$ feature matrix, where $J(x) = d_1(x) d_2(x)$, and apply $H_l$ framework discussed in Sections~\ref{ssec:h0},~\ref{ssec:h1},~\ref{ssec:h2}.
\end{enumerate}

\subsection{PCA-Reduced Fingerprints}
\label{ssec:red_fing}

In the case of higher-order fingerprints, such as $H_1, H_2$ (especially when applied to CNN features), the length, $n$, of each individual fingerprint may be much larger than the number of fingerprints, $N$, in the dataset, i.e. $n \gg N$.
If so, then PCA can be applied to the stack of fingerprints, $\bm{F} \in \mathbb{R}^{N \times n}$, to reduce from an $N \times n$ array to an $N \times N$ array.
This speeds up classifier training post-fingerprinting and compresses fingerprints, whilst retaining geometric information encoded into the features.

\section{Learning Tasks}
\label{sec:learning}

The ultimate goal of microstructural fingerprinting is to develop the tools to produce informative and compressed
representation of the microstructure associated with a particular population of microstructures.
This representation should be generic enough that fingerprints can be utilised for solution of a variety of problems, e.g. the development of machine learning models for classification and/or regression tasks.
In the present work, classification tasks are considered, where a category label assigned to each micrograph is predicted utilising the microstructural fingerprint as input.
Classification tasks can be split into three main categories, namely supervised learning (SL), semi-supervised learning (SSL) and unsupervised learning (UL).

Suppose that a fraction $p \in [0,1]$ of the data is labelled.
Then, SL is the case where $p=1$, UL has $p=0$ and SSL covers everything in between with $p \in (0, 1)$.
When $p \in (0,1)$, one could always discard the proportion $1-p$ of unlabelled data and proceed with SL,
which may often produce similar results, especially when $p$ and/or the original dataset is very large.
However, if a small proportion of data $p$ is labelled and a large amount of data is available,
it is prudent to leverage both labelled and unlabelled data towards the ultimate learning goal.
Labels can be propagated through the unlabelled images in the dataset and this forms the basis of SSL.

\subsection{Supervised Learning: $p=1$}
\label{ssec:sl}

A plethora of methods are widely available for SL.
Here, we compare support vector machine (SVM)~\cite{cortes1995support} with random forest (RF)~\cite{breiman_random_2001} approaches.

\subsubsection{Support Vector Machine}

SVM is one of the most widely used SL classifiers and has previously shown success for microstructure classification tasks~\cite{holm2}.
The aim is to find the optimal hyperplane for separating the data (the hyperplane with maximum margin, where the margin is defined as the distance between the hyperplane and the nearest data point).
In other words, we search for optimal values, $\bm{w}, b$, such that
\[
    \begin{cases}
        \bm{w}^\top \bm{F}_i - b \geq 1 - \xi_i, & \text{for } y_i = +1,\\
        \bm{w}^\top \bm{F}_i - b \leq -1 + \xi_i, & \text{for } y_i = -1.
    \end{cases}
\]
where $\bm{F}_i \equiv f(x_i)$ is the fingerprint corresponding to image $x_i$ for $i \in \{1, \dots, N\}$, given a set of $N$ images, $\bm{w}$ is a vector of weights, $b$ denotes a measure of bias and $y_i$ is a label assigned to $\bm{F}_i$ such that $y_i = +1$ if $\bm{F}_i$ is on one side of the hyperplane and $y_i = -1$ otherwise.
As the data may not be linearly separable, a slack variable $\xi_i$ is included for each training fingerprint to allow for some misclassification.
We then optimise by minimising
\[
    \|\bm{w}\|^2 + C \left( \sum_{i=1}^{n} \xi_i \right),
\]
where $C$ is a parameter to be specified and is typically set to 1, although cross-validation can be utilised to tune this.
The solution can then be given by
\[
    \bm{w} = \sum_i \alpha_i y_i \bm{F_i},
\]
where the $\alpha_i$ are to be determined numerically.
This can be done by considering the Lagrange dual problem and solving the Karush-Kuhn-Tucker (KKT) conditions.
Further details can be found in~\cite{bishop2006}.
We can then define class prediction, $g(x)$, for a test image, $x$, with fingerprint $\bm{F}$ as
\begin{equation} \label{eq:svm1}
g(x) = \text{sgn} \left( \sum_{i=1}^{n} \alpha_i y_i (\bm{F}_i \cdot \bm{F}) -b \right).
\end{equation}
Note that this is a binary classification, based on the sign of the solution.
This can be extended to multi-class problems and is discussed at the end of this section.

The input space (the space containing our fingerprints) may not be linearly separable.
To alleviate this, a map can be constructed to some feature space, $\phi(\bm{F}_i)$, which is selected to transform the input space into a space which is linearly separable.
Choosing the correct feature space can be challenging and an alternative approach that removes the need for such a map is to apply a suitable kernel.
If we choose our kernel, $K$, such that $K(\bm{F}_i, \bm{F}_j) = \phi(\bm{F}_i) \cdot \phi(\bm{F}_j)$,
we can rewrite our prediction~\eqref{eq:svm1} as
\[
    g(x) = \text{sgn} \left( \sum_{i=1}^{n} \alpha_i y_i K(\bm{F}_i, \bm{F}) -b \right),
\]
which is free of $\phi$.

A linear kernel results in essentially reverting back to solve equation~\eqref{eq:svm1}.
This works well for CNN fingerprints.
A $\chi^2$ kernel is utilised for VBOW, as recommended in~\cite{holm2}, however, this requires $\bm{F}_i \geq \bm{0}$ for all $i \in \{1, \dots, N\}$.
If $\bm{F}_i$ has been normalised in such a way that this condition is violated, a linear kernel should be applied.

As discussed above, SVM is a binary classifier, but there are several options to extend to multi-classification tasks by concatenating multiple binary classifiers.
Here we use a so-called error-correcting output code (ECOC)~\cite{dietterich1995}, which assigns a unique binary string to each class label.
The binary strings then allow for multiple binary classification problems, which can be combined to solve the multi-class problem.

\subsubsection{Random Forest}
\label{ssec:rf}

Random forests are an extension to decision tree classifiers.
A decision tree is a set of nodes and decisions, ultimately resulting in a prediction~\cite{breiman_classification_1998} .
They may be used for classification problems (as in our case), but they can also be applied to regression-based problems.
The root node at the top of the tree will contain all fingerprints.
The set of fingerprints is then split and propagated into a new pair of nodes.
Several iterations of this process form a tree.
At each node (including the root) a binary decision is made as to whether or not
\[
    \bm{F}_{i_j} < a,
\]
for some fixed value $a$ (selected during training to minimise a cost function), where $i \in \{1, \dots, N\}$ indexes the images and $j \in \{1, \dots, J(x_i)\}$ indexes the features in the $i^{\text{th}}$ image.

The Gini index, $\mathcal{G}$, is a measure of the expected error-rate at a given node~\cite{murphy2012} and is given by
\[
    \mathcal{G} = 1 - \sum_c \hat{\pi}_c^2,
\]
where $\hat{\pi}_c$ denotes the probability that a fingerprint drawn at random from the node will be of class $c$.
$\mathcal{G}$ is determined for each pair of nodes being fed and is combined as a weighted average across both nodes to provide the Gini impurity of the split.
The Gini criterion can then be used to prune the decision tree by removing splits with relatively large Gini impurity~\cite{bishop2006}.

Each decision splits the fingerprints into subsets, which are subsequently fed into other nodes until, finally, the nodes at the end of each branch contain only fingerprints from one class (or the specified maximum depth is reached).
These final nodes are known as leaf nodes.
Once trained, test data can be propagated through the decision tree to form a prediction, based on which leaf node the test data reaches.

Decision trees have a high variance.
To alleviate this, several decision trees can be trained and the ultimate prediction is given by the
prediction with the most votes, i.e. the answer which the majority of the decision trees arrived at (or the average prediction for regression-based tasks).
This is a process known as bootstrap aggregating (bagging)~\cite{breiman_bagging_1996}.
Bagging decision trees that are trained in this way can cause the decision trees to be correlated, due to the fact that the whole set of fingerprints received at each node is utilised.
This can lead to overfitting.
Random forest is essentially bagging, however,
during decision tree training, a random subset of fingerprints is utilised at each split (rather than the whole set).
This reduces correlation between decision trees, resulting in a random forest.

\subsection{Unsupervised Learning: $p=0$}
\label{ssec:ul}

For the UL task, we can use $k$-means clustering, as described already in
Section~\ref{sec:finger}, as well as spectral clustering~\cite{shi2000normalized}.
The latter is a graph-based method that requires suitable transformation of the image data before essentially applying $k$-means clustering to the constructed graph.

A graph is comprised of nodes and edges.
Each node denotes a fingerprint from an individual image and there are several options available for defining the edges/connections.
Here, we consider $k$ nearest neighbours~\cite{altman_introduction_1992} with $k = 10$, where neighbours are identified based on minimising Euclidean distance.
An adjacency matrix, $A \in \mathbb{R}^{J(x) \times J(x)}$, is constructed such that
\[
    \begin{cases}
        A_{ij} = 1, & \text{if } \bm{F}_j \text{ is a nearest neighbour of } \bm{F}_i,\\
        A_{ij} = 0, & \text{if } \bm{F}_j \text{ is not a nearest neighbour of } \bm{F}_i.
    \end{cases}
\]
Note the directionality of this behaviour. $A$ is not necessarily symmetric.

The degree matrix, $D \in \mathbb{R}^{J(x) \times J(x)}$, is also constructed.
This is a diagonal matrix such that $D_{ii}$ denotes the number of nodes connected to node $i$.
The Graph Laplacian, $L$, is then given by
\[
    L = D - A.
\]

Eigenvalues and eigenvectors are computed for $L$.
Eigenvalues can be used to estimate the optimal number of clusters (if not previously defined in the scope of the classification task) by plotting and searching for the first discontinuity in the plot.
This may be subjective at times, given the discrete nature of eigenvalues.
Finally, for a $k$-class problem with $N$ fingerprints, the first $k-1$ non-zero eigenvectors are calculated and form a $k \times N$ matrix.
Each row contains information about an optimal cut to make on the Graph Laplacian to cluster the fingerprints (nodes in the graph).
$k$-means clustering is performed on the eigenvectors, resulting in an $N$-dimensional vector of labels.
This serves as our classification for each of the $N$ images.

Labels assigned in this way do not directly correspond to class labels (in theory the true class labels don't even exist for UL).
However, this is useful if labels are not known and the aim is to group ``similar'' data together, rather than classify each group.
Regardless, we do have labels in this case and it is possible to construct a map from the predicted labels to the actual labels.
This enables us to quantify the accuracy of the unsupervised learning methods.

\subsection{Semi-Supervised Learning: $p \in (0,1)$}
\label{ssec:ssl}

For the SSL task, we again consider the Graph Laplacian, $L$, defined in Section~\ref{ssec:ul}.
However, now that we have some label information, we can look to propagate those labels through the graph, rather than seeking optimal cuts as in UL.
Here, we consider two methods for label propagation - Laplace learning and Poisson learning.
Jeff Calder's Python package, GraphLearning~\cite{jwcalder_graphlearning}, was used to test these learning methods.

\subsubsection{Laplace Learning}

Label propagation via Laplace learning was proposed in~\cite{Zhu2002LearningFL} and is based on solving Laplace's equation, i.e.
\[
    \begin{cases}
        \mathcal{L} u(\bm{F}_i) = 0, & \text{if $pN < i \leq N$},\\
        u(\bm{F}_i) = y_i, & \text{if $i \leq pN$},
    \end{cases}
\]
where $\mathcal{L} u(\bm{F}_i)$ denotes the Graph Laplacian, $L$, $p$ is the fraction of data that have labels associated to them and $N$ is the total number of fingerprints.

This is solved for $u(\bm{F}_i)$, such that $u(\bm{F}_i) = [u_1(\bm{F}_i), \dots, u_k(\bm{F}_i)]$ for the $k$-class problem.
The labels $y_i$, for $pN < i \leq N$, are then predicted as
\[
    y_i = \argmax_{j \in \{1, \dots, k\}}{u_j(\bm{F}_i)}.
\]

\subsubsection{Poisson Learning}

An alternative method for label propagation is presented in~\cite{calder2020poisson}, called Poisson learning.
It is proposed that this method offers a performance boost over Laplace learning at low label rates.

We define the average label vector, $\bar{y}$, as
\[
    \bar{y} = \frac{1}{pN} \sum_{j=1}^{pN} y_j
\]
and solve the Poisson equation
\[
    \mathcal{L} u(\bm{F}_i) = \sum_{j=1}^{pN} (y_j - \bar{y}) \delta_{ij},
\]
for $i = 1, \dots, N$, where
\[
    \delta_{ij} = \begin{cases}
                      1, & \text{if $i = j$,}\\
                      0, & \text{otherwise.}
    \end{cases}
\]
Note that no boundary conditions are required for Poisson learning: label information is instead encoded into a forcing term.

\section{Fingerprint Quality Assessment}
\label{sec:quality}

In a practical sense, the quality of a microstructural fingerprint is a balance between the amount of compression achieved and the amount of useful information that is retained.
To assess this, we must first define what we mean by useful information, but this is dependent on the intended use of the fingerprint.
In this case, we consider a classification task.

Classification tasks require the data to be split into a training set and a test set.
The test set will remain hidden from the model during training.
Following training, the test set is fed into the trained model and an accuracy score is computed.

It is important that several iterations of the model are implemented to minimise random error.
To this end, 10-fold cross-validation~\cite{Refaeilzadeh2009} is utilised.
This requires the data to be split into 10 distinct subsets (with even sampling across each microstructure class).
10 iterations of the classification task are then performed, with each subset utilised as the test set exactly once.
For each iteration, the remaining 9 subsets form the training set.

Following classification, the accuracy score for a task with $K$ classes is computed as
\[
    \sum_{i=1}^K C_{ii}/ \sum_{i,j=1}^K C_{ij},
\]
where $C \in \mathbb{R}^{K \times K}$ is the confusion matrix whose $i^{\text{th}}$ row indicates the true class and $j^{\text{th}}$ column indicates the machine prediction for the test data.
In particular, if no test data is mislabelled, then the accuracy score will be 1,
whereas, if all test data is mislabelled, then the accuracy score is 0.
The accuracy score is converted to a percentage accuracy for clarity in following sections.
Note that this will vary each time the simulation is run, due to the randomness of the train/test split.

In the case of the supervised and semi-supervised learning problems,
the confusion matrix is computed on the test data.
In the case of an unsupervised learning problem, predicted data labels are defined up to permutation of labels.
In other words, the data can be split into sets based on similarity, but the labels themselves carry no real meaning.
To compute an accuracy score in this case, the label mapping between actual labels and predicted labels which yields the highest accuracy is selected and the predicted labels are permuted accordingly.

\section{Data}\label{sec:data}

\subsection{Dataset 1: LFTi64}
\label{ssec:lfti64}

\begin{figure}[!ht]
    \centering
    \begin{subfigure}{0.4\textwidth}
        \centering
        \includegraphics[width=\textwidth]{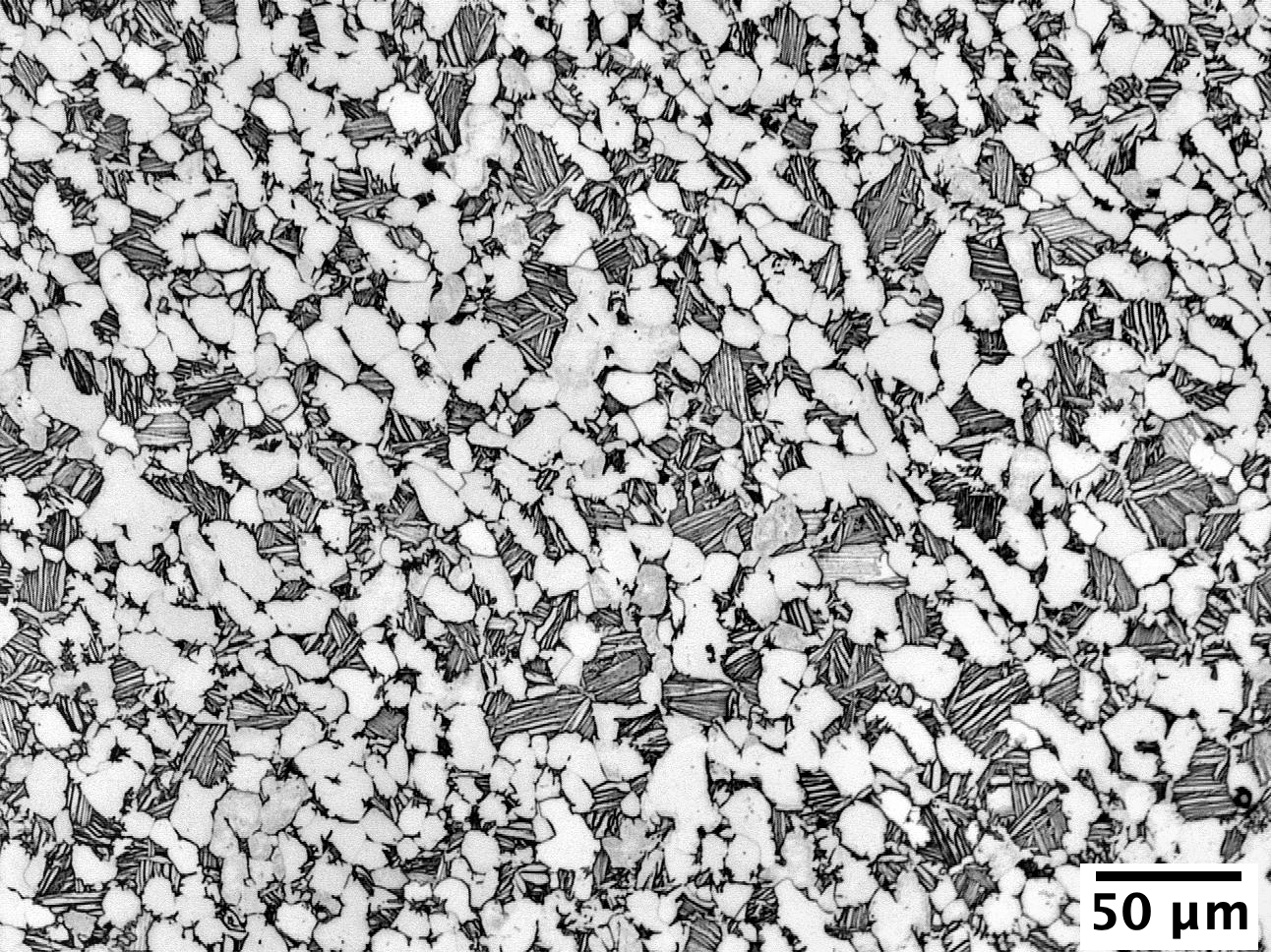}
        \caption{Bi-modal}
    \end{subfigure}
    \hspace{3em}
    \begin{subfigure}{0.4\textwidth}
        \centering
        \includegraphics[width=\textwidth]{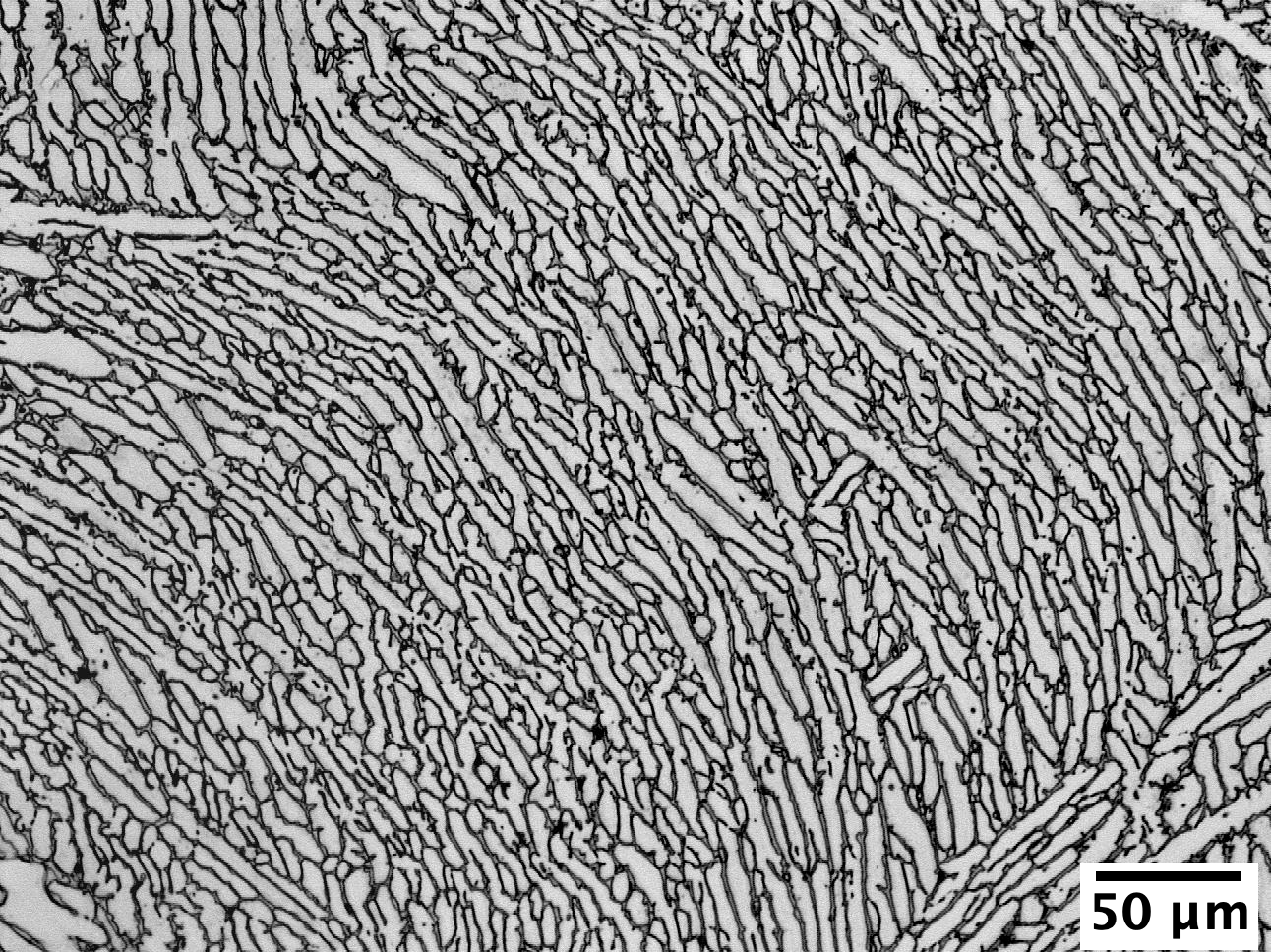}
        \caption{Lamellar}
    \end{subfigure}
    \caption{Example micrographs from LFTi64 dataset.}
    \label{fig:data_set_1}
\end{figure}

Dataset 1 (LFTi64) was curated within the LightForm group at the University of Manchester and is available on Zenodo~\cite{lfti64}.
The dataset consists of 40 optical micrographs of Titanium-6Al-4V.
20 images in the dataset are classified/labelled as a bi-modal equiaxed microstructure,
whilst the remaining 20 images are labelled as lamellar.
Figure~\ref{fig:data_set_1} provides some examples from the dataset.

The images in this dataset are relatively large (a small patch can be extracted as a representative volume element), benefit from homogeneous illumination and are completely devoid of any artifacts.
On the surface, this may seem ideal and, in some sense, this is true.
However, fingerprints constructed from this dataset were too distinct for each class to yield anything other than accuracies of 100\% across the board.
Hence, classification results generated from this dataset have been omitted from this report.

The main benefit to applying fingerprinting techniques to this dataset is the visual interpretability of the fingerprints themselves.
Values assigned to each cluster centre in the fingerprint may not be interpretable,
in the sense that we can't glean anything specific about the characteristics of the microstructure,
however, provided suitable ordering of the cluster centres, fingerprints from separate classes are clearly distinct by eye.
Also, fingerprints from micrographs within the same class are arguably ``similar'', at least by eye.
A distance metric could be employed to quantify this level of similarity.

\begin{figure}[!ht]
    \centering
    \begin{subfigure}[b]{0.4\textwidth}
        \centering
        \includegraphics[width=\textwidth]{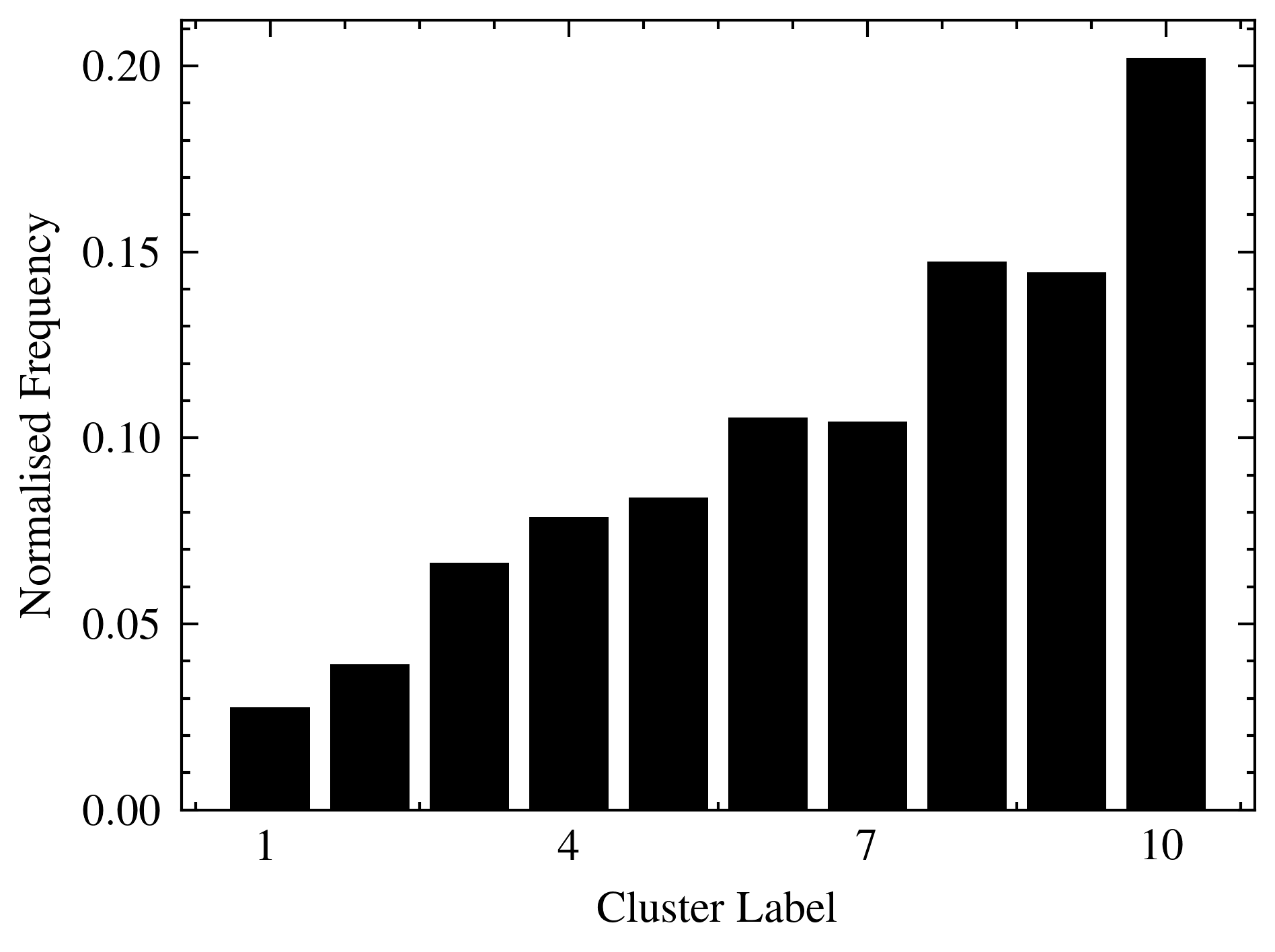}
        \caption{Bi-modal}
    \end{subfigure}%
    \hspace{3em}
    \begin{subfigure}[b]{0.4\textwidth}
        \centering
        \includegraphics[width=\textwidth]{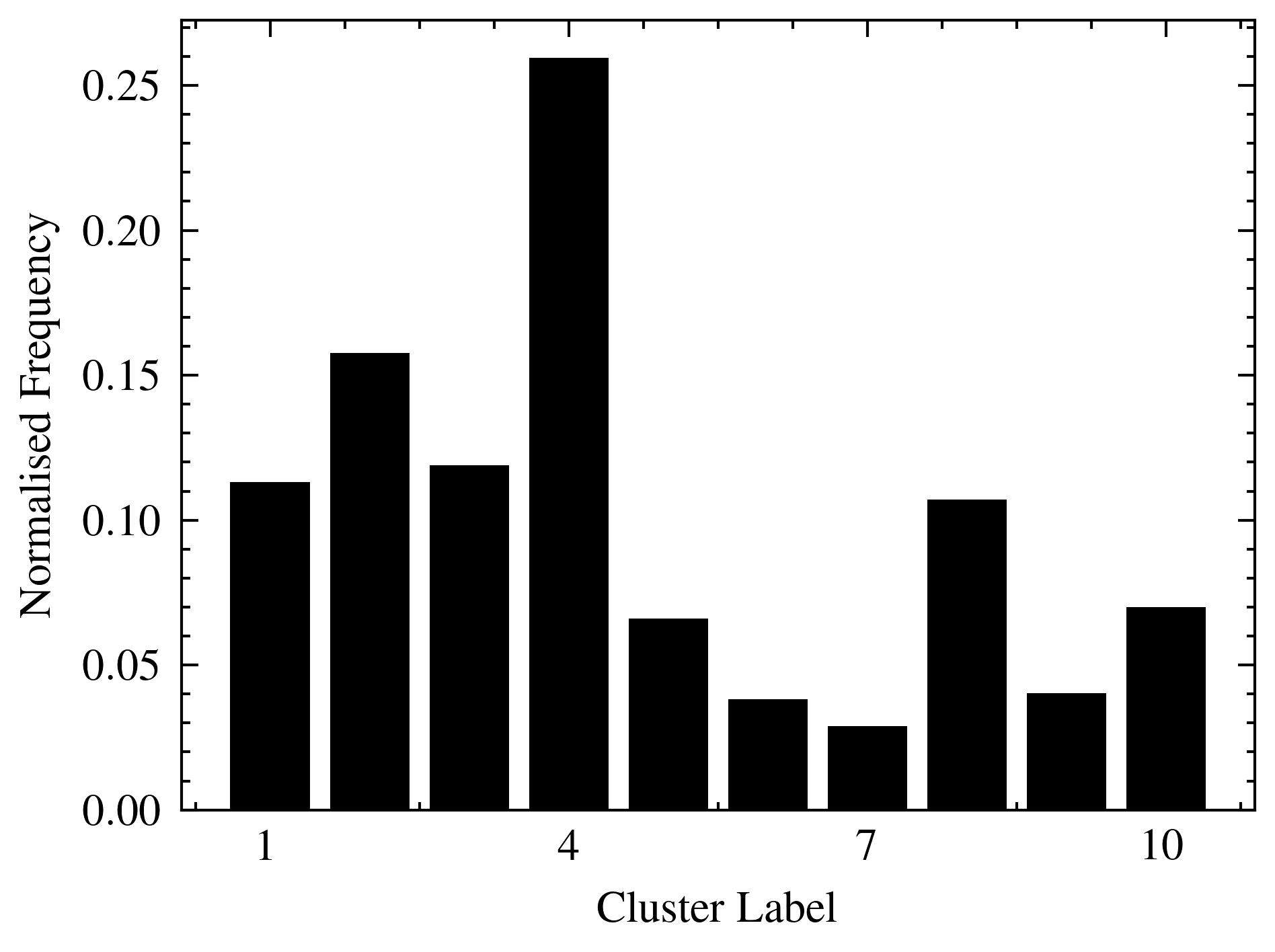}
        \caption{Lamellar}
    \end{subfigure}
    \begin{subfigure}[b]{0.4\textwidth}
        \centering
        \includegraphics[width=\textwidth]{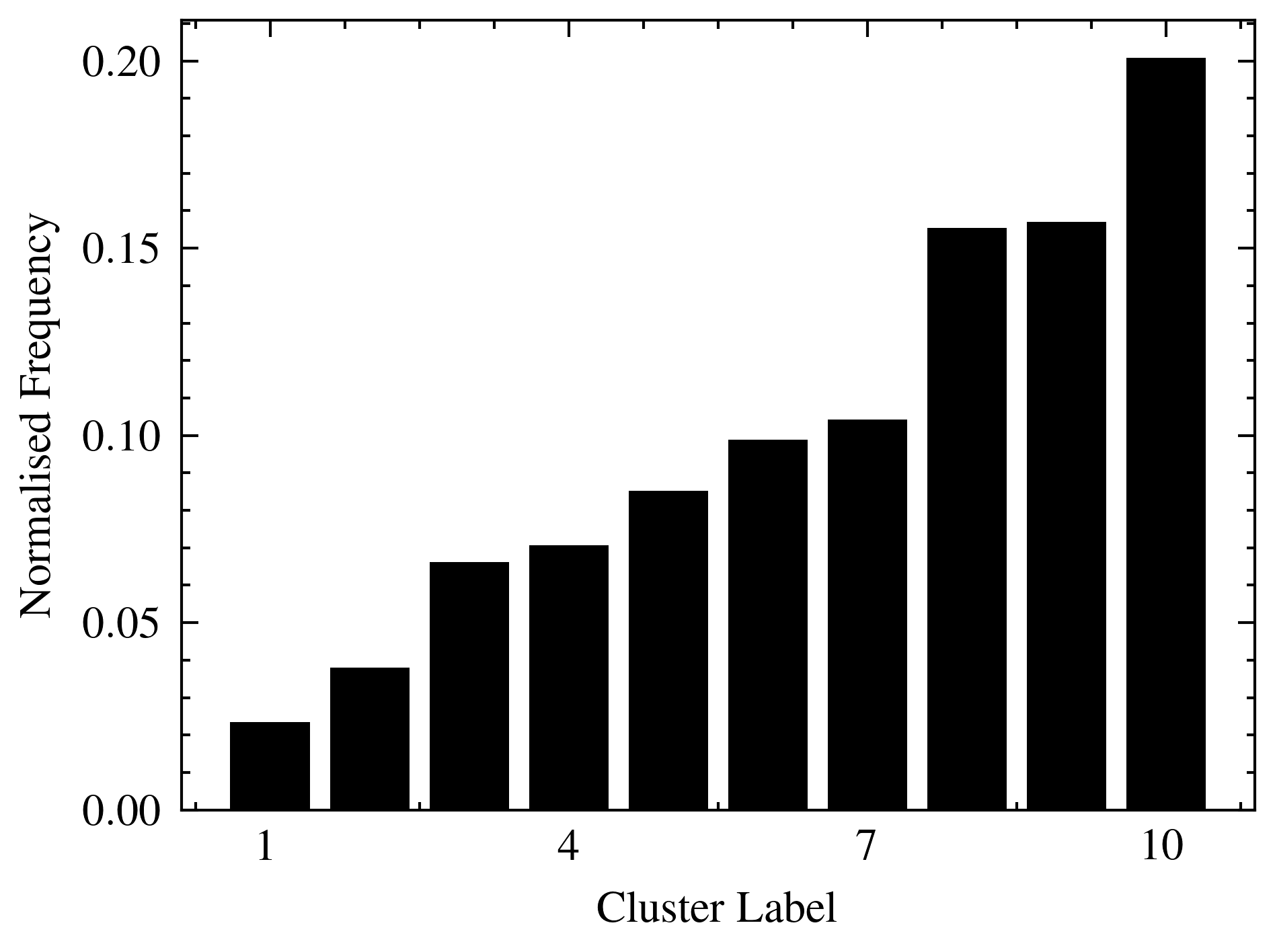}
        \caption{Bi-modal}
    \end{subfigure}%
    \hspace{3em}
    \begin{subfigure}[b]{0.4\textwidth}
        \centering
        \includegraphics[width=\textwidth]{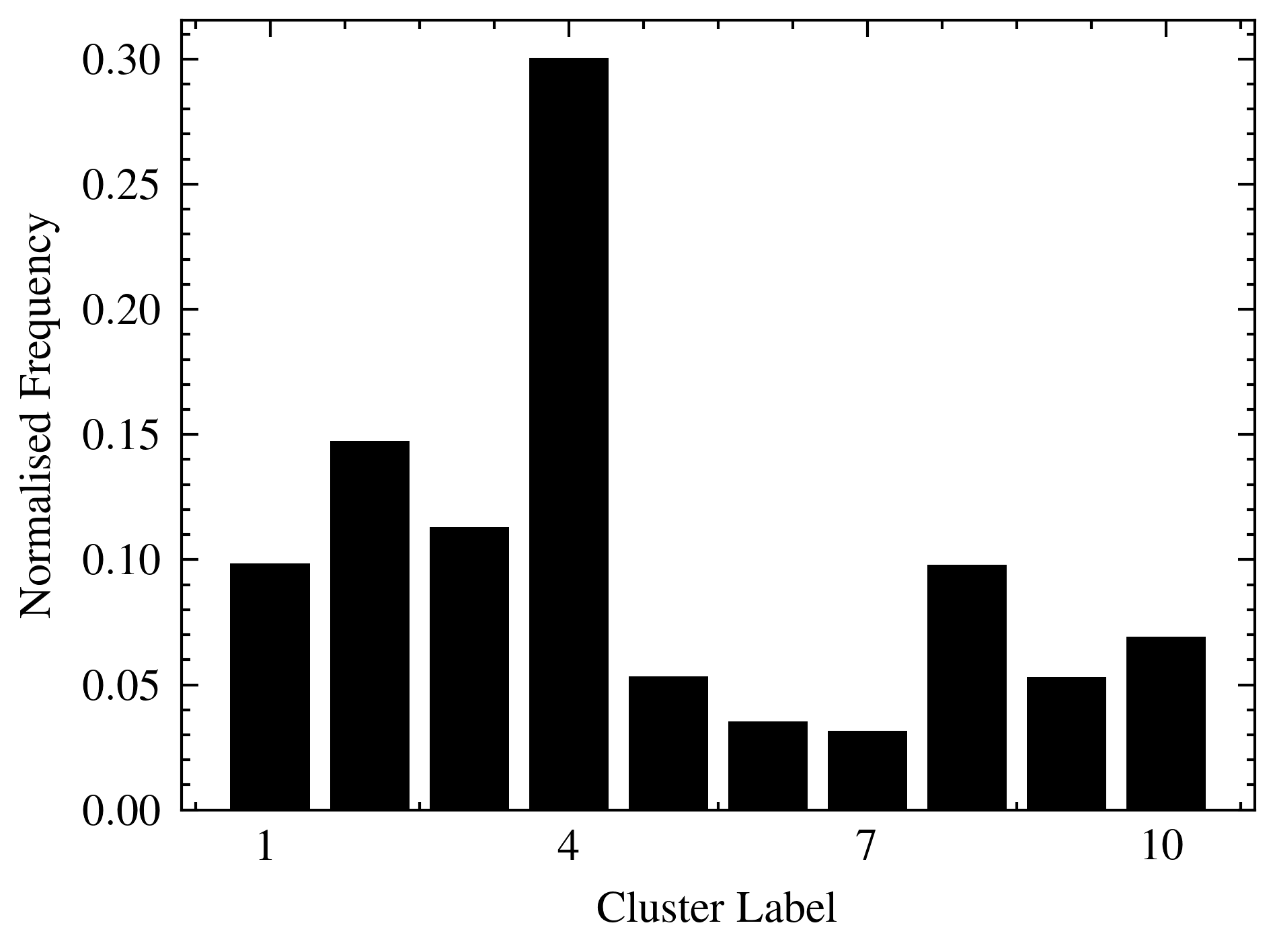}
        \caption{Lamellar}
    \end{subfigure}
    \caption{Example fingerprints from LFTi64 dataset, constructed from SURF features clustered with 10 cluster centres. These are ordered such that the mean of bi-modal microstructural fingerprints is in ascending order.}
    \label{fig:dataset1_fingers}
\end{figure}

Figure~\ref{fig:dataset1_fingers} shows some examples of SURF $H_{0, 10}$ fingerprints (zeroth order fingerprint with 10 cluster centres constructed from SURF keypoint features).
Cluster centres are ordered such that the mean fingerprint across all bi-modal micrographs is in ascending order.
Hence, bi-modal fingerprints shown here are approximately in ascending order, but not necessarily monotonically increasing.

\subsection{Dataset 2: UHCS600}
\label{ssec:uhcs600}

Dataset 2 was curated at Carnegie Mellon University and is presented in~\cite{decost_uhcsdb_2017}.
The dataset contains $961$ micrographs of ultrahigh carbon steel samples with seven different morphologies (listed in Table~\ref{tab:cmu-uhcs-classes}).
There are lots of variations in illumination/contrast and there is an abundance of defects/artifacts.
Hence, this is a much richer dataset for fingerprint quality assessment.

\begin{table}[!ht]
    \centering
    \begin{tabular}{ |c|c| }
        \hline
        \textbf{Morphology} & \textbf{Number of Micrographs} \\
        \hline
        Spheroidite                     & 373 \\
        \hline
        Carbide Network                 & 212 \\
        \hline
        Pearlite                        & 124 \\
        \hline
        Pearlite + Spheroidite          & 107 \\
        \hline
        Widmanst{\"a}tten Cementite     & 81 \\
        \hline
        Pearlite + Widmanst{\"a}tten    & 27 \\
        \hline
        Martensite / Bainite            & 36 \\
        \hline
    \end{tabular}
    \caption{CMU-UHCS Dataset Classes.}
    \label{tab:cmu-uhcs-classes}
\end{table}

\begin{figure}[!ht]
    \centering
    \begin{subfigure}[b]{0.3\textwidth}
        \centering
        \includegraphics[width=\textwidth]{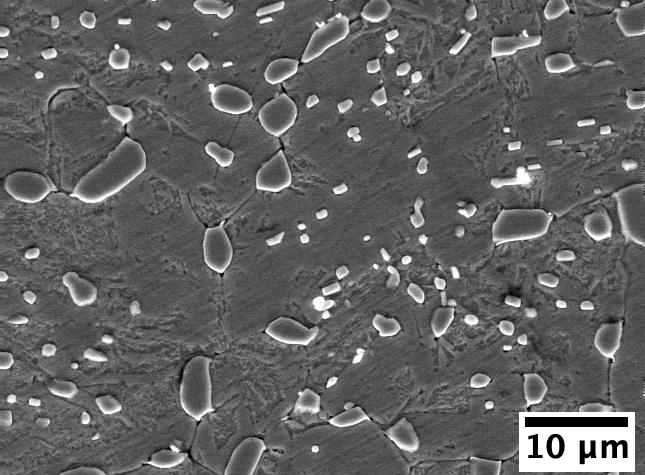}
        \caption{Spheroidite}
    \end{subfigure}
    \hfill
    \begin{subfigure}[b]{0.3\textwidth}
        \centering
        \includegraphics[width=\textwidth]{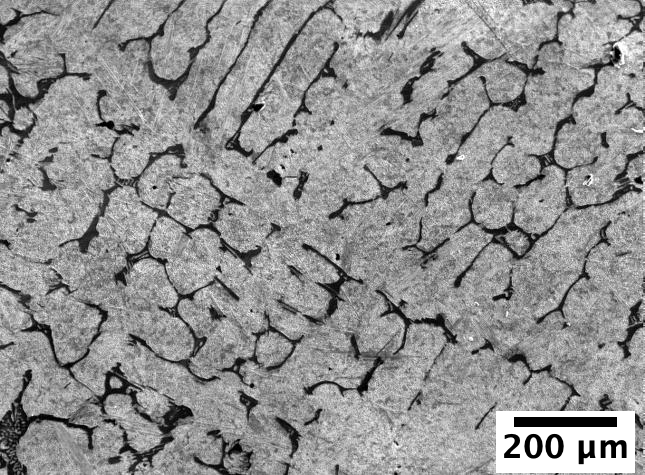}
        \caption{Carbide Network}
    \end{subfigure}
    \hfill
    \begin{subfigure}[b]{0.3\textwidth}
        \centering
        \includegraphics[width=\textwidth]{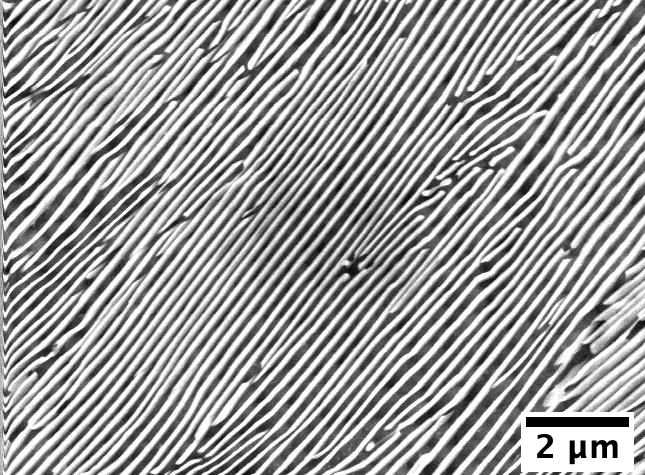}
        \caption{Pearlite}
    \end{subfigure}
    \caption{Example microstructures from UHCS600 dataset.}
    \label{fig:data_set_2}
\end{figure}

The dataset was reduced to 600 images across 3 morphologies by randomly sampling 200 spheroidite micrographs, 200 carbide network and 200 across pearlite and pearlite + spheroidite.
We will refer to the pearlite + spheroidite as just pearlite from now on.
It is worth noting that the presence of spheroidite in the pearlite micrographs has the potential to cause misclassification as spheroidite.
This does not cause too much of an issue, but, as mentioned later in Section~\ref{sec:discussion}, this attributes to a bottleneck in accuracy.
This reduced dataset is analogous to what was called UHCS600 in~\cite{holm2}.
However, the images are selected randomly
and so the realisation used here is likely different from that work and the results cannot be exactly/directly compared.
Examples of representative microstructures from each class are shown in Figure~\ref{fig:data_set_2}.

\section{Numerical Results}
\label{sec:numerics}

10-fold cross-validation accuracy scores were generated across various cluster sizes for fingerprints of order $l$, where $l = 0, 1, 2$, considering SIFT, SURF, AlexNet and VGG19 fingerprints.
The UHCS600 dataset was utilised to pose a 3-class classification task.
SVM and random forest were tested for solving the supervised learning task, $k$-means and spectral clustering for unsupervised learning and, finally, Laplace and Poisson learning for semi-supervised learning, with $p=0.05$ (labels provided for just 5\% of micrographs in the training set).
The supplementary Python package can be found on GitHub~\cite{mftools}.
Package requirements include Scikit-Learn~\cite{scikit-learn} for SL and UL classifiers, GraphLearning~\cite{jwcalder_graphlearning} for SSL classifiers and PyTorch~\cite{NEURIPS2019_9015} for CNN-based feature extraction.
Table~\ref{tab:parameters} provides all parameters used for testing.

\begin{table}[!ht]
    \centering
    \begin{tabular}{ |c|c|c| }
        \hline
        \textbf{Classifier}     & \textbf{Parameter}        & \textbf{Value}\\
        \hline
        \multirow{4}{*}{SVM}    & $C$                       & 1.0\\
        \cline{2-3}
        & $\gamma$                  & $1/J(x)$\\
        \cline{2-3}
        & \multirow{2}{*}{kernel}   & $\chi^2$ (VBOW)\\
        \cline{3-3}
        &                           & `linear' (CNN)\\
        \hline
        \multirow{3}{*}{RF}     & number of estimators      & 10000\\
        \cline{2-3}
        & max depth                 & 10\\
        \cline{2-3}
        & min no. elements per leaf & 1\\
        \hline
        $k$-means               & $k$ (number of classes)   & 3\\
        \hline
        \multirow{2}{*}{Spectral}   & $A$ (adjacency matrix)  & 10 nearest neighbours\\
        \cline{2-3}
        & $k$ (number of classes)                           & 3\\
        \hline
        \multirow{2}{*}{Laplace}    & $p$ (label rate)      & 0.05\\
        \cline{2-3}
        & $A$ (adjacency matrix)                            & 10 nearest neighbours\\
        \hline
        \multirow{2}{*}{Poisson}    & $p$ (label rate)      & 0.05\\
        \cline{2-3}
        & $A$ (adjacency matrix)                            & 10 nearest neighbours\\
        \hline
    \end{tabular}
    \caption{Table of parameters utilised to generate results.}
    \label{tab:parameters}
\end{table}

Tables~\ref{tab:sift-results},~\ref{tab:surf-results} and~\ref{tab:cnn-results}, which can be found in the appendix, show results for SIFT, SURF and CNN fingerprints, respectively.
Accuracy scores have been converted into percentage accuracy for clarity.
Standard deviation is not included in the tables to save space.
Regardless, standard deviation was consistent and within $O(0.01)$ across the board.
Highest accuracy scores for each learning method are emboldened to highlight the optimal fingerprint structure for solving each task.

In Table~\ref{tab:cnn-results}, subscripts ``F'' and ``MP'' refer to flattened and max pooled final convolution layers, respectively.
The superscript ``full'' refers to full-scale images (no transformation when preprocessing) and ``resize'' refers to input images having been deformed to meet the recommended input size of $224 \times 224$.
All cases where ``full'' or ``resize'' is not specified are to be regarded as ``full''.
Reduced dimensionality fingerprints are labelled with superscript ``$\text{red}\mathcal{P}$'' and ``$\text{red}F$'', corresponding to reduced feature population prior to fingerprint construction and reduced fingerprint dimensionality via PCA, respectively, and superscript ``diag'' denotes $H_2$ fingerprints reduced from $K \times d \times d$ to $K \times d$ via diagonalisation.

\section{Discussion}
\label{sec:discussion}

We have presented a framework for constructing fingerprints from microstructural image data, which applies to a host of feature extraction methods.
This is a data-driven approach to encoding microstructural information into a compressed form that is representative of the microstructure.
The fingerprint quality was assessed via a variety of classification tasks, but this could be extended to regression-based tasks where material processing parameters or mechanical properties are predicted, subject to a suitably labelled dataset.
It is not possible, however, to directly reconstruct the micrograph from the fingerprints in this case, although some options for fingerprint construction are being explored that would enable this, such as variational autoencoders (VAEs).

Generally, CNN-based fingerprints offer a performance boost over keypoint-based fingerprints.
However, the margin is relatively small.
For the SL task, the highest classification accuracy comes from applying $H_{1, 30}^\text{v}$ to the final convolution layer of VGG and employing SVM for classification, which gives a classification accuracy of $98.6\% \pm 1.44$.
This is in close agreement with the findings in~\cite{holm2}.
This accuracy score is matched by the same fingerprints with subsequent PCA reduction down to 600-dimensional vectors (VGG $H_{1, 30}^{\text{v, red}F}$), which has a classification accuracy of $98.6\% \pm 1.50$, and is closely followed by $H_{2, 10}^{\text{v, red}F}$ applied to AlexNet features, which has an accuracy of $98.4\% \pm 1.55$.
As well as generating useful descriptions of microstructure, one of the main aims for fingerprinting is to compress the information in the image as much as possible and, as such, application of PCA in this context is highly desirable.

It is important to note here the inherent randomness in the cross-validation train/test splits.
To ensure reliability of these scores, 100 iterations of the 10-fold cross-validation for SVM were computed and averaged.
The bottleneck in accuracy at around 98.5\% can be attributed to certain images in the dataset.
In fact, the same 6 images were misclassified 100\% of the time across these 100 iterations for VGG.
These were either images from the ``pearlite + spheroidite'' set that were given a true class label of ``pearlite'', but actually consisted of mainly spheroidite and were consistently misclassified as such;
or they were relatively low-magnification images from the spheroidite set that were consistently misclassified as pearlite due to individual grains not being resolved at the low magnification and appearing to merge, resembling lamellae.
The other SL classification method tested (random forest) deals well with lower-dimensional fingerprints, but breaks down slightly for high-dimensional fingerprints and becomes very computationally expensive, unlike SVM.

Image preprocessing is crucial to obtaining such high accuracies.
If the input image is resized via deformation, the cross-validation accuracy drops (in our case, from $96.7\% \pm 2.24$ to $92.2\% \pm 2.89$ for max pooled AlexNet).
Therefore, CNN results in Table~\ref{tab:cnn-results} (unless labelled ``resize'') were generated using full-scale input images and resizing to match the recommended input size is best avoided.
This is perfectly fine, as we are only interested in utilising the feature extraction portion of the pretrained CNNs in this case.

As suggested in~\cite{calder2020poisson}, Poisson learning offers an improvement over Laplace learning for SSL at low label rates ($p = 0.05$), in most cases.
Simply max pooling the final convolution layer provides the highest classification accuracy for SSL ($91.3\% \pm 2.48$ for AlexNet and $94.0\% \pm 0.98$ for VGG).
This is very close to the SL classification accuracy for the same fingerprinting method in each case ($96.0\% \pm 2.38$ and $97.7\% \pm 1.70$, respectively).
One of the largest difficulties currently faced in machine learning for materials science is in the curation and distribution of sufficiently large datasets.
There are several decades-worth of microstructural image data that have been collected,
but the majority are not suitably labelled and it is a laborious and costly task to label such massive amounts of data.
The fact that SSL performs so well with just $5\%$ of the data being labelled suggests that perhaps we do not need to label data on such a large scale, but just a handful of representative images may be sufficient.

For the UL task, VBOW fingerprints offer the highest classification accuracy, with a maximum of $83.5\% \pm 5.98$ achieved by applying spectral clustering to SIFT $H_{0, 50}$ fingerprints.
Spectral clustering outperformed $k$-means almost entirely across the board.
The upshot here is that graph-based methods can provide an incredibly useful representation of micrograph data.

Increasing the order of the $H_l$ fingerprints from 0 to 1 can boost performance.
Centring and $L_2$-normalising (VLAD) can offer a further performance boost.
Generally, $H_l$ appears to reach the point of diminishing returns at $H_2$, in this context.
This may be due to the fact that the dataset is relatively small and the $H_2$ fingerprints are high-dimensional.
However, $H_2$ does offer a performance boost when applying SVM to AlexNet fingerprints (provided suitable PCA reduction) and, also, for spectral clustering applied to VGG fingerprints for UL and when Poisson learning applied to SIFT fingerprints for SSL.
$H_2$ is computationally expensive, though, so
to reduce the computational cost and enable $H_2$ for CNN-based fingerprints, either the diagonal was taken from each $K \times d \times d$ fingerprint to form a $K \times d$ fingerprint (results with the superscript ``diag''), or PCA was applied in two different ways.
Primarily, feature arrays generated from each image individually were reduced prior to feature population construction, not in dimensionality, but in the number of features, by transposing the feature array before and after applying PCA.
This resulted in classification accuracies approaching the highest performing methods applied to the whole feature population ($98.2\% \pm 1.89$ for $H_{2, 20}^{\text{v}}$ fingerprints constructed from reduced population of VGG features compared to maximum of $98.6\% \pm 1.44$ for non-reduced feature population).
Performing $H_1^\text{v}$ on the same reduced features results in a drop in accuracy and so $H_2^\text{v}$ should be considered if vast amounts of features are extracted.
Secondly, PCA reduction of the fingerprint space to reduce dimensionality (as opposed to reduction of the number of features present in the population $\mathcal{P}$) can offer a performance boost for the SL task, as discussed above, but this enhancement seems to be restricted to SL in this setting and fails to yield results much better than random for SSL and UL tasks.

SIFT was shown to outperform SURF overall (the highest accuracy achieved by SIFT of $98.0\% \pm 1.70$ with $H_1^\text{v}$ compared to $97.7\% \pm 1.33$ for SURF).
However, this is a relatively small difference and the added compression afforded by SURF, alongside the reduced run time, could be considered favourable in this case.
It takes 336 seconds for SIFT dictionary creation from all 600 input images, compared to 237 seconds for SURF.
This has a knock-on effect to clustering and, therefore, fingerprint construction (584 seconds for $k$-means clustering SIFT features with $k=20$ and only 305 seconds for SURF).
This difference would become more significant for larger datasets and larger values of $k$.

Introducing the multi-scale VBOW fingerprints can offer a further improvement in cross-validation accuracy score, whilst maintaining compression.
For example, $H_{0, 20+30+50}$ outperforms $H_{0, 100}$ for random forest applied to SURF-based fingerprints (with classification accuracies of $94.7\% \pm 3.79$ and $93.8\% \pm 2.52$, respectively), but both are 100-dimensional representations of the input images.

\begin{figure}[!ht]
    \centering
    \begin{subfigure}[b]{0.4\textwidth}
        \centering
        \includegraphics[width=\textwidth]{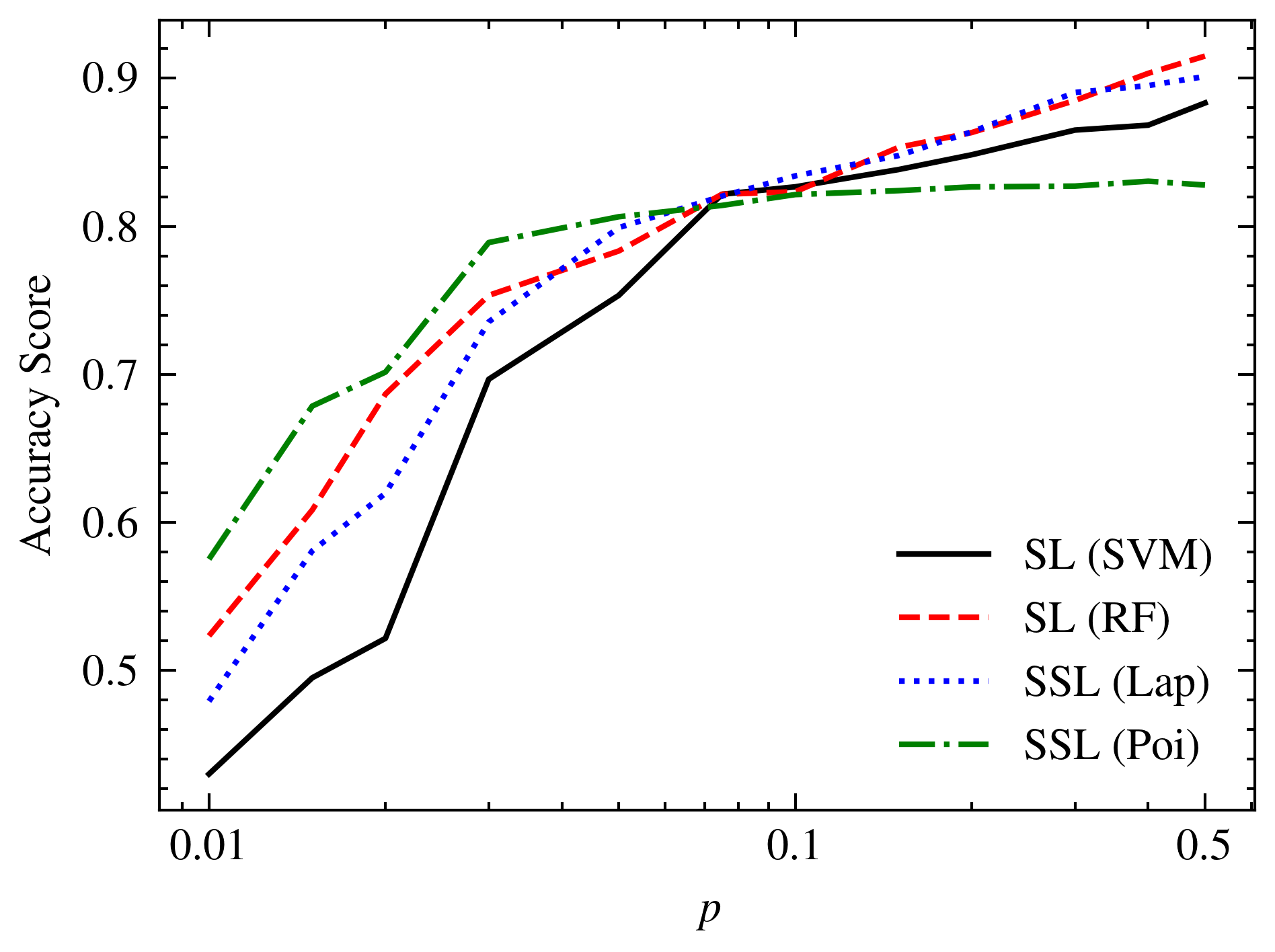}
        \caption{SURF $H_{0, 20}$}
    \end{subfigure}
    \hspace{3em}
    \begin{subfigure}[b]{0.4\textwidth}
        \centering
        \includegraphics[width=\textwidth]{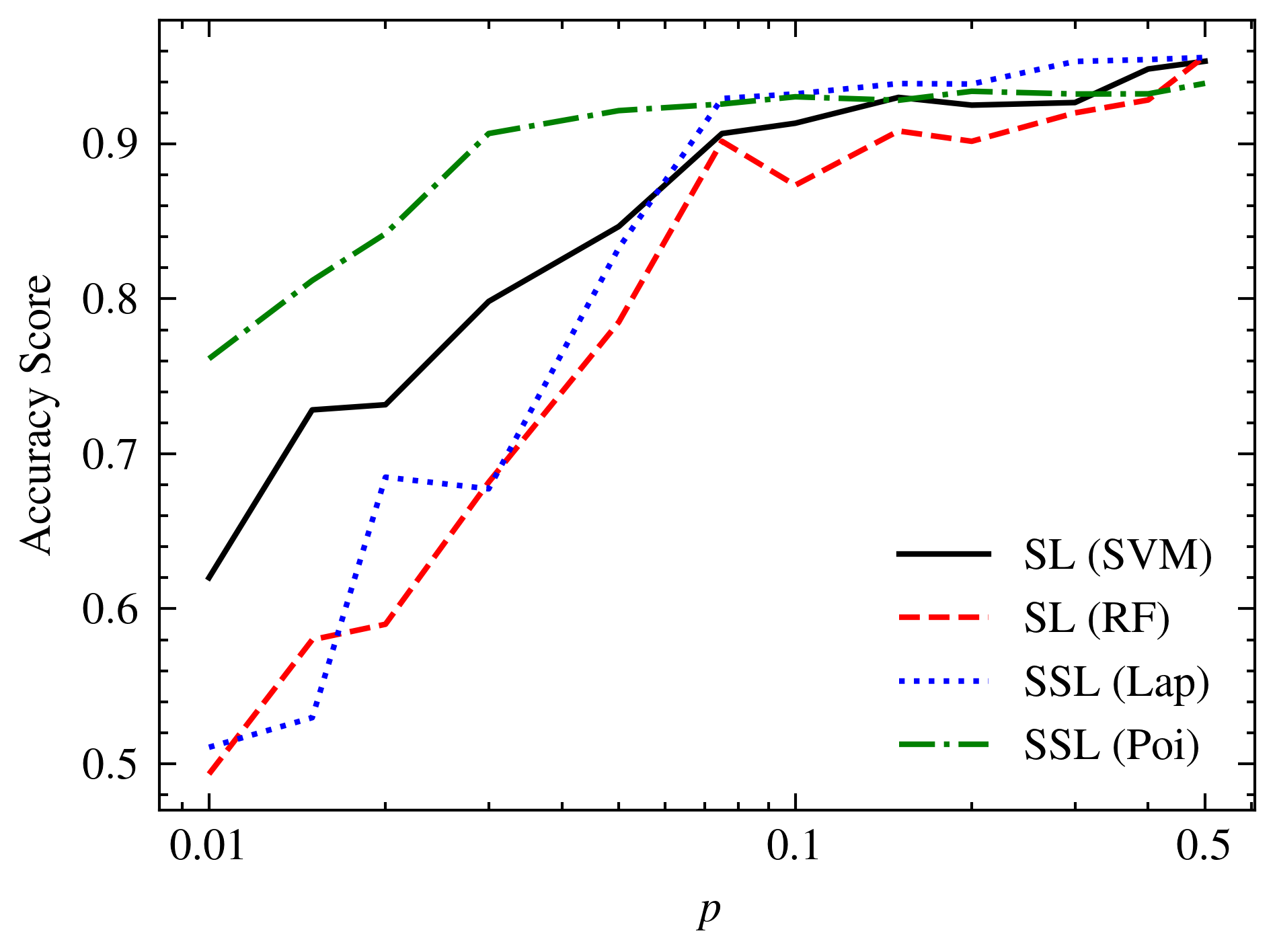}
        \caption{AlexNet max pooled}
    \end{subfigure}
    \caption{Reduced training set of size $pN$ for SL compared with SSL label propagation with $pN$ labels, where $N$ is the total number of images (600 in this case) and $0.01 \leq p \leq 0.5$ denotes the fraction of data that have labels.}
    \label{fig:sl_ssl}
\end{figure}

Reducing the training dataset size for SL to $p$ times its initial size allows us to quantify the performance boost offered by label propagation via SSL over training an SL classifier on available labels and discarding unlabelled data.
Figure~\ref{fig:sl_ssl} provides plots of the classification accuracy for SVM, RF, Laplace learning and Poisson learning applied to SURF $H_{0, 20}$ and max pooled AlexNet features.
Both plots confirm a performance boost offered by label propagation at low label rates.
This is most apparent with the AlexNet max pooled fingerprints, but both methods benefit from label propagation via Poisson learning for $p \leq 0.1$.
This value is likely dependent on the size of the dataset in question.
Poisson learning does offer an improvement over Laplace learning for low label rates (as proposed in~\cite{calder2020poisson}), but Laplace learning overtakes at approximately $p = 0.1$, alongside the SL methods.

\section{Conclusions}
\label{sec:conclusions}

We have compared various options for image-based feature extraction on microstructural image data, namely SIFT, SURF and transfer learning from two pretrained CNNs (AlexNet and VGG), pretrained on the ImageNet dataset.
We have also proposed a statistical framework for constructing fingerprints from these features, which incorporates some classical computer vision approaches (VBOW and VLAD).
The quality of these fingerprints was assessed via a range classification tasks - supervised learning (SL), semi-supervised learning (SSL) and unsupervised learning (UL).
The main conclusions are summarised below.

\begin{itemize}
    \item Semi-supervised learning (SSL) with less than 10\% of data being assigned labels ($0 < p < 0.1$) can outperform supervised learning (SL) trained on an equivalent subset of the training data with unlabelled data discarded, i.e.\ label propagation to unlabelled training data is more beneficial than discarding unlabelled data and performing SL\@.
    \item SSL on a dataset with 5\% of labels for the training set can approach accuracy of SL with a maximum classification accuracy of 94.0\% for Poisson learning on max-pooled VGG features, compared to 98.6\% for SVM applied to VGG $H_1^{\text{v}}$ fingerprints.
    \item For transfer learning with pretrained networks, VGG19 offers a maximum improvement of 0.6\% accuracy over AlexNet via SVM classification, however, the dimensionality of outputs from VGG are 2.72 times larger that those of AlexNet and subsequent clustering of features is NP-hard.
    \item Centring and normalising fingerprints generally offers a further performance boost.
    For high-throughput scenarios where time is valuable, AlexNet could be favourable.
    \item Spectral clustering offers a performance-boost over $k$-means clustering for UL in almost all cases, with an average improvement of 9.1\% across all fingerprinting methods.
    \item UL benefits from low-dimensional fingerprints and performs best on zeroth-order fingerprints constructed from SIFT features, with a maximum classification accuracy of 81.3\% on the dataset considered here.
\end{itemize}

We have explored a variety of approaches to generating feature vectors for microstructural images and developed a framework for combining these into microstructural fingerprints (concise descriptions of microstructural image data that capture intrinsic properties of the microstructure).
We have explored the relative merits of these different approaches by considering a simple classification task.
Clear next steps in this research would be to extend the validation and use of these fingerprints to regression-based tasks suitable for exploring process-structure-property relationships and quantitatively assessing the difference between similar microstructures via some form of distance metric between fingerprints.
We are also investigating additional methods for fingerprint construction, such as variational autoencoders (VAEs) and vision transformers, as tools for encoding microstructural image data.

All methods for fingerprint construction discussed can, in theory, be applied to 3-channel colour images and, as such, could be applied to EBSD data;
either in the form of orientation distribution functions (ODFs), or from spatial reconstructions.
This should also be investigated, as it will enable richer microstructural information to be encoded in the fingerprints, which could be leveraged to construct more robust process-structure-property relationships than afforded by greyscale image data.

\section{Acknowledgements}
\label{sec:acknowledgements}

MDW was supported by the University of Manchester and Rolls-Royce plc under the Engineering and
Physical Sciences Research Council (EPSRC) grant EP/S022635/1 and Science Foundation Ireland (SFI) grant 18/EPSRC-CDT/3584.
AT and KJHL were supported by The Alan Turing Institute under the EPSRC grant EP/N510129/1.
CPR was supported by a University Research Fellowship of the Royal Society.
PJW was supported by the Henry Royce Institute for Advanced Materials, funded through EPSRC grants EP/R00661X/1, EP/S019367/1, EP/P025021/1 and EP/P025498/1.

We would like to thank Matt Thorpe for discussions regarding SSL and for pointing us to Poisson learning and the GraphLearning Python package used for that.
We would also like to thank Ben Saunders and Christopher Collins for their helpful discussions throughout the whole research process.
Computational work was performed on the Computational Shared Facility of the University of Manchester.

\bibliographystyle{elsarticle-num}
\bibliography{references}

\begin{thebibliography}{10}
\expandafter\ifx\csname url\endcsname\relax
  \def\url#1{\texttt{#1}}\fi
\expandafter\ifx\csname urlprefix\endcsname\relax\def\urlprefix{URL }\fi
\expandafter\ifx\csname href\endcsname\relax
  \def\href#1#2{#2} \def\path#1{#1}\fi

\bibitem{mftools}
M.~White, A.~Tarakanov,
  \href{https://github.com/mikedwhite/microstructural-fingerprinting-tools}{Microstructural
  {Fingerprinting} {Tools}} (2021).
\newline\urlprefix\url{https://github.com/mikedwhite/microstructural-fingerprinting-tools}

\bibitem{kalidindi2015hierarchical}
S.~R. Kalidindi, Hierarchical materials informatics: novel analytics for
  materials data, Elsevier, 2015.
\newblock \href {https://doi.org/10.1016/C2012-0-07337-1}
  {\path{doi:10.1016/C2012-0-07337-1}}.

\bibitem{agrawal2016perspective}
A.~Agrawal, A.~Choudhary, Perspective: Materials informatics and big data:
  Realization of the fourth paradigm of science in materials science, APL
  Materials 4~(5) (2016).
\newblock \href {https://doi.org/10.1063/1.4946894}
  {\path{doi:10.1063/1.4946894}}.

\bibitem{holm4}
E.~A. Holm, R.~Cohn, N.~Gao, A.~R. Kitahara, T.~P. Matson, B.~Lei, S.~R.
  Yarasi, Overview: Computer vision and machine learning for microstructural
  characterization and analysis, Metallurgical and Materials Transactions A 51
  (2020) 5985--5999.
\newblock \href {https://doi.org/10.1007/s11661-020-06008-4}
  {\path{doi:10.1007/s11661-020-06008-4}}.

\bibitem{molkeri2022}
A.~Molkeri, D.~Khatamsaz, R.~Couperthwaite, J.~James, R.~Arróyave, D.~Allaire,
  A.~Srivastava, On the importance of microstructure information in materials
  design: Psp vs pp, Acta Materialia 223 (2022) 117471.
\newblock \href {https://doi.org/10.1016/j.actamat.2021.117471}
  {\path{doi:10.1016/j.actamat.2021.117471}}.

\bibitem{schmitz2016}
G.~J. Schmitz, B.~Böttger, M.~Apel, J.~Eiken, G.~Laschet, R.~Altenfeld,
  R.~Berger, G.~Boussinot, A.~Viardin, Towards a metadata scheme for the
  description of materials – the description of microstructures, Science and
  Technology of Advanced Materials 17~(1) (2016) 410--430.
\newblock \href {https://doi.org/10.1080/14686996.2016.1194166}
  {\path{doi:10.1080/14686996.2016.1194166}}.

\bibitem{xie2021}
Q.~Xie, M.~Suvarna, J.~Li, X.~Zhu, J.~Cai, X.~Wang, Online prediction of
  mechanical properties of hot rolled steel plate using machine learning,
  Materials \& Design 197 (2021) 109201.
\newblock \href {https://doi.org/10.1016/j.matdes.2020.109201}
  {\path{doi:10.1016/j.matdes.2020.109201}}.

\bibitem{ford2021}
E.~Ford, K.~Maneparambil, S.~Rajan, N.~Neithalath, Machine learning-based
  accelerated property prediction of two-phase materials using microstructural
  descriptors and finite element analysis, Computational Materials Science 191
  (2021) 110328.
\newblock \href {https://doi.org/10.1016/j.commatsci.2021.110328}
  {\path{doi:10.1016/j.commatsci.2021.110328}}.

\bibitem{yucel2020}
B.~Yucel, S.~Yucel, A.~Ray, L.~Duprez, S.~R. Kalidindi, Mining the
  {Correlations} {Between} {Optical} {Micrographs} and {Mechanical}
  {Properties} of {Cold}-{Rolled} {HSLA} {Steels} {Using} {Machine} {Learning}
  {Approaches}, Integrating Materials and Manufacturing Innovation 9~(3) (2020)
  240--256.
\newblock \href {https://doi.org/10.1007/s40192-020-00183-3}
  {\path{doi:10.1007/s40192-020-00183-3}}.

\bibitem{yamanaka2020}
A.~Yamanaka, R.~Kamijyo, K.~Koenuma, I.~Watanabe, T.~Kuwabara, Deep neural
  network approach to estimate biaxial stress-strain curves of sheet metals,
  Materials \& Design 195 (2020) 108970.
\newblock \href {https://doi.org/10.1016/j.matdes.2020.108970}
  {\path{doi:10.1016/j.matdes.2020.108970}}.

\bibitem{ma2020image}
W.~Ma, E.~J. Kautz, A.~Baskaran, A.~Chowdhury, V.~Joshi, B.~Yener, D.~J. Lewis,
  Image-driven discriminative and generative machine learning algorithms for
  establishing microstructure-processing relationships, Journal of Applied
  Physics 128~(13) (2020) 134901.
\newblock \href {https://doi.org/10.1063/5.0013720}
  {\path{doi:10.1063/5.0013720}}.

\bibitem{lambardo2020}
A.~Gayon-Lombardo, L.~Mosser, N.~P. Brandon, S.~J. Cooper, Pores for thought:
  generative adversarial networks for stochastic reconstruction of 3d
  multi-phase electrode microstructures with periodic boundaries, npj
  Computational Materials 6 (2020).
\newblock \href {https://doi.org/10.1038/s41524-020-0340-7}
  {\path{doi:10.1038/s41524-020-0340-7}}.

\bibitem{withers2019}
T.~Burnett, P.~Withers, Completing the picture through correlative
  characterization, Nature Materials 18 (2019) 1041--1049.
\newblock \href {https://doi.org/10.1038/s41563-019-0402-8}
  {\path{doi:10.1038/s41563-019-0402-8}}.

\bibitem{holm1}
B.~L. DeCost, E.~A. Holm, A computer vision approach for automated analysis and
  classification of microstructural image data, Computational Materials Science
  110 (2015) 126--133.
\newblock \href {https://doi.org/10.1016/j.commatsci.2015.08.011}
  {\path{doi:10.1016/j.commatsci.2015.08.011}}.

\bibitem{holm2}
B.~L. DeCost, T.~Francis, E.~A. Holm, Exploring the microstructure manifold:
  image texture representations applied to ultrahigh carbon steel
  microstructures, Acta Materialia 133 (2017) 30--40.
\newblock \href {https://doi.org/10.1016/j.actamat.2017.05.014}
  {\path{doi:10.1016/j.actamat.2017.05.014}}.

\bibitem{holmdnn}
B.~L. DeCost, B.~Lei, T.~Francis, E.~A. Holm, High throughput quantitative
  metallography for complex microstructures using deep learning: a case study
  in ultrahigh carbon steel, Microscopy and Microanalysis 25~(1) (2019) 21--29.
\newblock \href {https://doi.org/10.1017/S1431927618015635}
  {\path{doi:10.1017/S1431927618015635}}.

\bibitem{azimi2018advanced}
S.~M. Azimi, D.~Britz, M.~Engstler, M.~Fritz, F.~M{\"u}cklich, Advanced steel
  microstructural classification by deep learning methods, Scientific Reports
  8~(1) (2018) 1--14.
\newblock \href {https://doi.org/10.1038/s41598-018-20037-5}
  {\path{doi:10.1038/s41598-018-20037-5}}.

\bibitem{holm3}
A.~R. Kitahara, E.~A. Holm, Microstructure cluster analysis with transfer
  learning and unsupervised learning, Integrating Materials and Manufacturing
  Innovation 7~(3) (2018) 148--156.
\newblock \href {https://doi.org/10.1007/s40192-018-0116-9}
  {\path{doi:10.1007/s40192-018-0116-9}}.

\bibitem{mueller2021machine}
M.~Mueller, D.~Britz, F.~Muecklich, Machine learning for microstructure
  classification: How to assign ground truth in the most objective way,
  Advanced Materials \& Processes 179~(1) (2021) 16--21.

\bibitem{ssl1}
C.~Kunselman, S.~Sheikh, M.~Mikkelsen, V.~Attari, R.~Arr{\'o}yave,
  Microstructure classification in the unsupervised context, Acta Materialia
  (2020).
\newblock \href {https://doi.org/10.2139/ssrn.3683591}
  {\path{doi:10.2139/ssrn.3683591}}.

\bibitem{ssl2}
A.~Tran, T.~Rodgers, T.~M. Wildey, Reification of latent microstructures: On
  supervised unsupervised and semi-supervised deep learning applications for
  microstructures in materials informatics, Tech. rep., Sandia National
  Lab.(SNL-NM), Albuquerque, NM (United States) (2020).
\newblock \href {https://doi.org/10.2172/1673174} {\path{doi:10.2172/1673174}}.

\bibitem{ssl3}
K.~Guo, M.~J. Buehler, A semi-supervised approach to architected materials
  design using graph neural networks, Extreme Mechanics Letters 41 (2020)
  101029.
\newblock \href {https://doi.org/10.1016/j.eml.2020.101029}
  {\path{doi:10.1016/j.eml.2020.101029}}.

\bibitem{ssl4}
C.~Kunselman, V.~Attari, L.~McClenny, U.~Braga-Neto, R.~Arroyave,
  Semi-supervised learning approaches to class assignment in ambiguous
  microstructures, Acta Materialia 188 (2020) 49--62.
\newblock \href {https://doi.org/10.2139/ssrn.3499091}
  {\path{doi:10.2139/ssrn.3499091}}.

\bibitem{filliat2007visual}
D.~Filliat, A visual bag of words method for interactive qualitative
  localization and mapping, in: Proceedings 2007 IEEE International Conference
  on Robotics and Automation, IEEE, 2007, pp. 3921--3926.
\newblock \href {https://doi.org/10.1109/ROBOT.2007.364080}
  {\path{doi:10.1109/ROBOT.2007.364080}}.

\bibitem{csurka2004visual}
G.~Csurka, C.~Dance, L.~Fan, J.~Willamowski, C.~Bray, Visual categorization
  with bags of keypoints, in: Workshop on statistical learning in computer
  vision, ECCV, Vol.~1, Prague, 2004, pp. 1--22.

\bibitem{lazebnik2005sparse}
S.~Lazebnik, C.~Schmid, J.~Ponce, A sparse texture representation using local
  affine regions, IEEE transactions on pattern analysis and machine
  intelligence 27~(8) (2005) 1265--1278.
\newblock \href {https://doi.org/10.1109/TPAMI.2005.151}
  {\path{doi:10.1109/TPAMI.2005.151}}.

\bibitem{nowak2006sampling}
E.~Nowak, F.~Jurie, B.~Triggs, Sampling strategies for bag-of-features image
  classification, in: European conference on computer vision, Springer, 2006,
  pp. 490--503.
\newblock \href {https://doi.org/10.1007/11744085_38}
  {\path{doi:10.1007/11744085_38}}.

\bibitem{jegou2010aggregating}
H.~J{\'e}gou, M.~Douze, C.~Schmid, P.~P{\'e}rez, Aggregating local descriptors
  into a compact image representation, in: 2010 IEEE computer society
  conference on computer vision and pattern recognition, IEEE, 2010, pp.
  3304--3311.
\newblock \href {https://doi.org/10.1109/CVPR.2010.5540039}
  {\path{doi:10.1109/CVPR.2010.5540039}}.

\bibitem{delhumeau2013revisiting}
J.~Delhumeau, P.-H. Gosselin, H.~J{\'e}gou, P.~P{\'e}rez, Revisiting the vlad
  image representation, in: Proceedings of the 21st ACM international
  conference on Multimedia, 2013, pp. 653--656.
\newblock \href {https://doi.org/10.1145/2502081.2502171}
  {\path{doi:10.1145/2502081.2502171}}.

\bibitem{cimpoi2015deep}
M.~Cimpoi, S.~Maji, A.~Vedaldi, Deep filter banks for texture recognition and
  segmentation, in: Proceedings of the IEEE conference on computer vision and
  pattern recognition, 2015, pp. 3828--3836.
\newblock \href {https://doi.org/10.1109/CVPR.2015.7299007}
  {\path{doi:10.1109/CVPR.2015.7299007}}.

\bibitem{jia2014caffe}
Y.~Jia, E.~Shelhamer, J.~Donahue, S.~Karayev, J.~Long, R.~Girshick,
  S.~Guadarrama, T.~Darrell, Caffe: Convolutional architecture for fast feature
  embedding, in: Proceedings of the 22nd ACM international conference on
  Multimedia, 2014, pp. 675--678.
\newblock \href {https://doi.org/10.1145/2647868.2654889}
  {\path{doi:10.1145/2647868.2654889}}.

\bibitem{parkhi2014compact}
O.~M. Parkhi, K.~Simonyan, A.~Vedaldi, A.~Zisserman, A compact and
  discriminative face track descriptor, in: Proceedings of the IEEE Conference
  on Computer Vision and Pattern Recognition, 2014, pp. 1693--1700.
\newblock \href {https://doi.org/10.1109/CVPR.2014.219}
  {\path{doi:10.1109/CVPR.2014.219}}.

\bibitem{bertozzi2012diffuse}
A.~L. Bertozzi, A.~Flenner, Diffuse interface models on graphs for
  classification of high dimensional data, Multiscale Modeling \& Simulation
  10~(3) (2012) 1090--1118.
\newblock \href {https://doi.org/10.1137/11083109X}
  {\path{doi:10.1137/11083109X}}.

\bibitem{calder2020poisson}
J.~Calder, B.~Cook, M.~Thorpe, D.~Slepcev, Poisson learning: Graph based
  semi-supervised learning at very low label rates, in: International
  Conference on Machine Learning, PMLR, 2020, pp. 1306--1316.

\bibitem{lowe1999object}
D.~G. Lowe, Object recognition from local scale-invariant features, in:
  Proceedings of the seventh IEEE international conference on computer vision,
  Vol.~2, IEEE, 1999, pp. 1150--1157.
\newblock \href {https://doi.org/10.1109/ICCV.1999.790410}
  {\path{doi:10.1109/ICCV.1999.790410}}.

\bibitem{lowe2004distinctive}
D.~G. Lowe, Distinctive image features from scale-invariant keypoints,
  International Journal of Computer Vision 60~(2) (2004) 91--110.
\newblock \href {https://doi.org/10.1023/B:VISI.0000029664.99615.94}
  {\path{doi:10.1023/B:VISI.0000029664.99615.94}}.

\bibitem{bay2008speeded}
H.~Bay, A.~Ess, T.~Tuytelaars, L.~Van~Gool, Speeded-up robust features
  ({S}{U}{R}{F}), Computer Vision and Image Understanding 110~(3) (2008)
  346--359.
\newblock \href {https://doi.org/10.1016/j.cviu.2007.09.014}
  {\path{doi:10.1016/j.cviu.2007.09.014}}.

\bibitem{gong2014multi}
Y.~Gong, L.~Wang, R.~Guo, S.~Lazebnik, Multi-scale orderless pooling of deep
  convolutional activation features, in: European conference on computer
  vision, Springer, 2014, pp. 392--407.
\newblock \href {https://doi.org/10.1007/978-3-319-10584-0_26}
  {\path{doi:10.1007/978-3-319-10584-0_26}}.

\bibitem{vgg}
K.~Simonyan, A.~Zisserman, Very deep convolutional networks for large-scale
  image recognition, arXiv:1409.1556 (2014).

\bibitem{shi2000normalized}
J.~Shi, J.~Malik, Normalized cuts and image segmentation, IEEE Transactions on
  pattern analysis and machine intelligence 22~(8) (2000) 888--905.
\newblock \href {https://doi.org/10.1109/34.868688}
  {\path{doi:10.1109/34.868688}}.

\bibitem{von2007tutorial}
U.~Von~Luxburg, A tutorial on spectral clustering, Statistics and Computing
  17~(4) (2007) 395--416.
\newblock \href {https://doi.org/10.1007/s11222-007-9033-z}
  {\path{doi:10.1007/s11222-007-9033-z}}.

\bibitem{sammut_encyclopedia_2010}
C.~Sammut, G.~I. Webb (Eds.), Encyclopedia of {Machine} {Learning}, Springer
  US, Boston, MA, 2010.
\newblock \href {https://doi.org/10.1007/978-0-387-30164-8}
  {\path{doi:10.1007/978-0-387-30164-8}}.

\bibitem{pourkamali-anaraki_scalable_2020}
F.~Pourkamali-Anaraki, Scalable spectral clustering with {Nyström}
  approximation: Practical and theoretical aspects, IEEE Open Journal of Signal
  Processing 1 (2020) 242--256.
\newblock \href {https://doi.org/10.1109/OJSP.2020.3039330}
  {\path{doi:10.1109/OJSP.2020.3039330}}.

\bibitem{van_fleet_haar_2011}
P.~J. Van~Fleet, The {Haar} {Wavelet} {Transformation}, in: Discrete {Wavelet}
  {Transformations}, John Wiley \& Sons, Inc., Hoboken, NJ, USA, 2011, pp.
  157--221.
\newblock \href {https://doi.org/10.1002/9781118032404.ch6}
  {\path{doi:10.1002/9781118032404.ch6}}.

\bibitem{ILSVRC15}
O.~Russakovsky, J.~Deng, H.~Su, J.~Krause, S.~Satheesh, S.~Ma, Z.~Huang,
  A.~Karpathy, A.~Khosla, M.~Bernstein, A.~C. Berg, L.~Fei-Fei, {ImageNet Large
  Scale Visual Recognition Challenge}, International Journal of Computer Vision
  (IJCV) 115~(3) (2015) 211--252.
\newblock \href {https://doi.org/10.1007/s11263-015-0816-y}
  {\path{doi:10.1007/s11263-015-0816-y}}.

\bibitem{alexnet}
A.~Krizhevsky, I.~Sutskever, G.~E. Hinton, Imagenet classification with deep
  convolutional neural networks, Advances in Neural Information Processing
  Systems 25 (2012) 1097--1105.

\bibitem{deng2009imagenet}
J.~Deng, W.~Dong, R.~Socher, L.-J. Li, K.~Li, L.~Fei-Fei, Imagenet: A
  large-scale hierarchical image database, in: 2009 IEEE conference on computer
  vision and pattern recognition, Ieee, 2009, pp. 248--255.
\newblock \href {https://doi.org/10.1109/CVPR.2009.5206848}
  {\path{doi:10.1109/CVPR.2009.5206848}}.

\bibitem{chatfield2014a}
K.~Chatfield, K.~Simonyan, A.~Vedaldi, A.~Zisserman, Return of the devil in the
  details: delving deep into convolutional nets, British Machine Vision
  Association, 2014, pp. 1--12.
\newblock \href {https://doi.org/10.5244/C.28.6} {\path{doi:10.5244/C.28.6}}.

\bibitem{cortes1995support}
C.~Cortes, V.~Vapnik, Support-vector networks, Machine Learning 20~(3) (1995)
  273--297.
\newblock \href {https://doi.org/10.1007/BF00994018}
  {\path{doi:10.1007/BF00994018}}.

\bibitem{breiman_random_2001}
L.~Breiman, Random forests, Machine Learning 45~(1) (2001) 5--32.
\newblock \href {https://doi.org/10.1023/A:1010933404324}
  {\path{doi:10.1023/A:1010933404324}}.

\bibitem{bishop2006}
C.~M. Bishop, Pattern Recognition and Machine Learning, Springer-Verlag,
  Berlin, Heidelberg, 2006.
\newblock \href {https://doi.org/10.5555/1162264} {\path{doi:10.5555/1162264}}.

\bibitem{dietterich1995}
T.~G. Dietterich, G.~Bakiri, Solving multiclass learning problems via
  error-correcting output codes, Journal of Artificial Intelligence Ressearch
  2~(1) (1995) 263–286.
\newblock \href {https://doi.org/10.5555/1622826.1622834}
  {\path{doi:10.5555/1622826.1622834}}.

\bibitem{breiman_classification_1998}
L.~Breiman (Ed.), Classification and regression trees, 1st Edition, Chapman \&
  Hall/CRC, Boca Raton, Fla., 1998.

\bibitem{murphy2012}
K.~P. Murphy, Machine Learning: A Probabilistic Perspective, The MIT Press,
  2012.
\newblock \href {https://doi.org/10.5555/2380985} {\path{doi:10.5555/2380985}}.

\bibitem{breiman_bagging_1996}
L.~Breiman, Bagging predictors, Machine Learning 24~(2) (1996) 123--140.
\newblock \href {https://doi.org/10.1007/BF00058655}
  {\path{doi:10.1007/BF00058655}}.

\bibitem{altman_introduction_1992}
N.~S. Altman, An introduction to kernel and nearest-neighbor nonparametric
  regression, The American Statistician 46~(3) (1992) 175.
\newblock \href {https://doi.org/10.2307/2685209} {\path{doi:10.2307/2685209}}.

\bibitem{jwcalder_graphlearning}
J.~Calder, M.~Rios~Flores, B.~Cook, M.~Jacobs, M.~Ettehad,
  \href{https://github.com/jwcalder/GraphLearning}{Graph{Learning}} (2020).
\newline\urlprefix\url{https://github.com/jwcalder/GraphLearning}

\bibitem{Zhu2002LearningFL}
X.~Zhu, Z.~Ghahramani, Learning from labeled and unlabeled data with label
  propagation, 2002.

\bibitem{Refaeilzadeh2009}
P.~Refaeilzadeh, L.~Tang, H.~Liu, Cross-Validation, Springer US, Boston, MA,
  2009, pp. 532--538.
\newblock \href {https://doi.org/10.1007/978-0-387-39940-9_565}
  {\path{doi:10.1007/978-0-387-39940-9_565}}.

\bibitem{lfti64}
M.~White, C.~Daniel, J.~Quinta~da Fonseca, X.~Zeng, N.~Byers, B.~Karnasiewicz,
  Lamellar and bi-modal ti-64 microstructure images (Nov. 2021).
\newblock \href {https://doi.org/10.5281/zenodo.5714384}
  {\path{doi:10.5281/zenodo.5714384}}.

\bibitem{decost_uhcsdb_2017}
B.~L. DeCost, M.~D. Hecht, T.~Francis, B.~A. Webler, Y.~N. Picard, E.~A. Holm,
  {UHCSDB}: {UltraHigh} {Carbon} {Steel} {Micrograph} {DataBase}: {Tools} for
  {Exploring} {Large} {Heterogeneous} {Microstructure} {Datasets}, Integrating
  Materials and Manufacturing Innovation 6~(2) (2017) 197--205.
\newblock \href {https://doi.org/10.1007/s40192-017-0097-0}
  {\path{doi:10.1007/s40192-017-0097-0}}.

\bibitem{scikit-learn}
F.~Pedregosa, G.~Varoquaux, A.~Gramfort, V.~Michel, B.~Thirion, O.~Grisel,
  M.~Blondel, P.~Prettenhofer, R.~Weiss, V.~Dubourg, J.~Vanderplas, A.~Passos,
  D.~Cournapeau, M.~Brucher, M.~Perrot, E.~Duchesnay, Scikit-learn: Machine
  learning in {P}ython, Journal of Machine Learning Research 12 (2011)
  2825--2830.

\bibitem{NEURIPS2019_9015}
A.~Paszke, S.~Gross, F.~Massa, A.~Lerer, J.~Bradbury, G.~Chanan, T.~Killeen,
  Z.~Lin, N.~Gimelshein, L.~Antiga, A.~Desmaison, A.~Kopf, E.~Yang, Z.~DeVito,
  M.~Raison, A.~Tejani, S.~Chilamkurthy, B.~Steiner, L.~Fang, J.~Bai,
  S.~Chintala, Pytorch: An imperative style, high-performance deep learning
  library, in: H.~Wallach, H.~Larochelle, A.~Beygelzimer, F.~d\textquotesingle
  Alch\'{e}-Buc, E.~Fox, R.~Garnett (Eds.), Advances in Neural Information
  Processing Systems 32, Curran Associates, Inc., 2019, pp. 8024--8035.

\end{thebibliography}

\newpage
\appendix
\section{Results Tables}

\begin{table}[!ht]
    \centering
    \resizebox{\textwidth}{!}{%
        \begin{tabular}{ |c|c|c|c|c|c|c| }
            \hline
            \multirow{2}{*}{\textbf{Method}} & \multicolumn{2}{c|}{\textbf{SL ($p=1$)}} & \multicolumn{2}{c|}{\textbf{UL ($p=0$)}} & \multicolumn{2}{c|}{\textbf{SSL ($p=0.05$)}} \\ \cline{2-7}
            & \textbf{SVM} & \textbf{RF} & \textbf{$k$-means} & \textbf{Spectral} & \textbf{Laplace} & \textbf{Poisson} \\
            \specialrule{.08em}{0em}{0em}
            $\text{SIFT} H_{0, 10}$              & $84.3$ & $91.7$ & $70.5$ & $78.5$ & $83.2$ & $81.8$     \\
            \hline
            $\text{SIFT} H_{0, 20}$              & $92.3$ & $95.2$ & $72.8$ & $80.2$ & $84.1$ & $84.0$     \\
            \hline
            $\text{SIFT} H_{0, 50}$              & $94.8$ & $96.3$ & $78.2$ & $\bm{83.5}$ & $88.8$ & $88.5$     \\
            \hline
            $\text{SIFT} H_{0, 100}$             & $94.5$ & $95.2$ & $\bm{81.3}$ & $82.7$ & $84.6$ & $87.6$     \\
            \hline
            $\text{SIFT} H_{0, 20 + 30}$         & $94.5$ & $95.2$ & $72.0$ & $79.2$ & $84.9$ & $85.3$    \\
            \hline
            $\text{SIFT} H_{0, 20 + 30 + 50}$    & $94.8$ & $96.5$ & $74.8$ & $81.3$ & $84.9$ & $86.9$    \\
            \specialrule{.08em}{0em}{0em}
            $\text{SIFT} H_{1, 10}$              & $96.3$ & $95.3$ & $54.2$ & $79.2$ & $86.2$ & $90.3$     \\
            \hline
            $\text{SIFT} H_{1, 50}$              & $97.0$ & $93.3$ & $50.3$ & $66.0$ & $73.4$ & $81.4$     \\
            \hline
            $\text{SIFT} H_{1, 10}^{\text{v}}$   & $96.8$ & $\bm{96.3}$ & $68.3$ & $69.8$ & $81.5$ & $87.8$     \\
            \hline
            $\text{SIFT} H_{1, 50}^{\text{v}}$   & $\bm{98.0}$ & $95.3$ & $73.5$ & $77.8$ & $85.6$ & $88.9$     \\
            \specialrule{.08em}{0em}{0em}
            $\text{SIFT} H_{2, 10}$              & $95.0$ & $95.0$ & $61.7$ & $75.5$ & $\bm{89.2}$ & $90.9$     \\
            \hline
            $\text{SIFT} H_{2, 20}$              & $97.2$ & $96.0$ & $64.0$ & $80.2$ & $83.7$ & $88.2$     \\
            \hline
            $\text{SIFT} H_{2, 10}^{\text{v}}$   & $96.8$ & $96.0$ & $53.2$ & $68.0$ & $86.9$ & $\bm{91.0}$     \\
            \hline
            $\text{SIFT} H_{2, 20}^{\text{v}}$   & $96.3$ & $\bm{96.3}$ & $54.2$ & $67.8$ & $84.3$ & $88.3$     \\
            \hline
        \end{tabular}}
    \caption{Percentage accuracy for classification with SIFT fingerprints as input, with highest accuracies emphasised in bold for each classification method.}
    \label{tab:sift-results}
\end{table}

\begin{table}[!ht]
    \centering
    \resizebox{\textwidth}{!}{%
        \begin{tabular}{ |c|c|c|c|c|c|c| }
            \hline
            \multirow{2}{*}{\textbf{Method}} & \multicolumn{2}{c|}{\textbf{SL ($p=1$)}} & \multicolumn{2}{c|}{\textbf{UL ($p=0$)}} & \multicolumn{2}{c|}{\textbf{SSL ($p=0.05$)}} \\ \cline{2-7}
            & \textbf{SVM} & \textbf{RF} & \textbf{$k$-means} & \textbf{Spectral} & \textbf{Laplace} & \textbf{Poisson} \\
            \specialrule{.08em}{0em}{0em}
            $\text{SURF} H_{0, 10}$              & $88.2$ & $91.7$ & $70.0$ & $74.8$ & $76.8$ & $79.4$     \\
            \hline
            $\text{SURF} H_{0, 20}$              & $89.5$ & $93.0$ & $73.8$ & $78.3$ & $81.5$ & $82.3$     \\
            \hline
            $\text{SURF} H_{0, 50}$              & $92.8$ & $93.8$ & $72.3$ & $75.5$ & $81.6$ & $75.6$     \\
            \hline
            $\text{SURF} H_{0, 100}$             & $94.8$ & $93.8$ & $74.5$ & $78.0$ & $81.7$ & $85.2$     \\
            \hline
            $\text{SURF} H_{0, 256}$             & $94.2$ & $92.0$ & $75.2$ & $73.5$ & $76.6$ & $72.0$     \\
            \hline
            $\text{SURF} H_{0, 20 + 30}$         & $93.5$ & $92.8$ & $76.3$ & $79.5$ & $83.1$ & $84.2$    \\
            \hline
            $\text{SURF}_{0, 20 + 30 + 50}$      & $94.3$ & $94.7$ & $72.5$ & $77.3$ & $84.1$ & $84.8$    \\
            \specialrule{.08em}{0em}{0em}
            $\text{SURF} H_{1, 20}$              & $93.2$ & $93.5$ & $65.7$ & $77.8$ & $77.4$ & $82.6$     \\
            \hline
            $\text{SURF} H_{1, 50}$              & $93.3$ & $94.8$ & $51.5$ & $76.3$ & $72.1$ & $83.8$     \\
            \hline
            $\text{SURF} H_{1, 100}$             & $94.2$ & $92.5$ & $48.5$ & $79.0$ & $52.4$ & $77.0$     \\
            \hline
            $\text{SURF} H_{1, 20}^{\text{v}}$   & $95.6$ & $\bm{96.3}$ & $77.2$ & $77.8$ & $87.9$ & $85.3$     \\
            \hline
            $\text{SURF} H_{1, 50}^{\text{v}}$   & $97.3$ & $95.7$ & $80.0$ & $80.3$ & $\bm{88.0}$ & $87.7$     \\
            \hline
            $\text{SURF} H_{1, 100}^{\text{v}}$  & $\bm{97.7}$ & $95.5$ & $\bm{80.5}$ & $\bm{81.8}$ & $85.5$ & $\bm{87.9}$     \\
            \specialrule{.08em}{0em}{0em}
            $\text{SURF} H_{2, 10}$              & $90.0$ & $94.5$ & $61.5$ & $72.8$ & $79.6$ & $83.7$     \\
            \hline
            $\text{SURF} H_{2, 20}$              & $93.3$ & $94.0$ & $58.5$ & $66.5$ & $73.8$ & $82.7$     \\
            \hline
            $\text{SURF} H_{2, 10}^{\text{v}}$   & $93.7$ & $93.3$ & $50.3$ & $66.0$ & $73.4$ & $81.4$     \\
            \hline
            $\text{SURF} H_{2, 20}^{\text{v}}$   & $96.4$ & $93.3$ & $47.2$ & $64.0$ & $39.1$ & $62.6$     \\
            \hline
        \end{tabular}}
    \caption{Percentage accuracy for classification with SURF fingerprints as input, with highest accuracies emphasised in bold for each classification method.}
    \label{tab:surf-results}
\end{table}

\begin{table}[!ht]
    \centering
    \resizebox{\textwidth}{!}{%
        \begin{tabular}{ |c|c|c|c|c|c|c| }
            \hline
            \multirow{2}{*}{\textbf{Method}} & \multicolumn{2}{c|}{\textbf{SL ($p=1$)}} & \multicolumn{2}{c|}{\textbf{UL ($p=0$)}} & \multicolumn{2}{c|}{\textbf{SSL ($p=0.05$)}} \\ \cline{2-7}
            & \textbf{SVM} & \textbf{RF} & \textbf{$k$-means} & \textbf{Spectral} & \textbf{Laplace} & \textbf{Poisson} \\
            \specialrule{.08em}{0em}{0em}
            $\text{AlexNet}_{\text{F}}^{\text{full}}$       & $95.5$ & $92.2$ & $38.0$ & $55.8$ & $35.9$ & $57.1$     \\
            \hline
            $\text{AlexNet}_{\text{MP}}^{\text{full}}$      & $96.7$ & $95.0$ & $60.8$ & $60.8$ & $83.7$ & $91.3$     \\
            \hline
            $\text{AlexNet}_{\text{F}}^{\text{resize}}$     & $92.7$ & $89.2$ & $35.5$ & $51.5$ & $45.3$ & $70.5$     \\
            \hline
            $\text{AlexNet}_{\text{MP}}^{\text{resize}}$    & $92.2$ & $92.8$ & $61.5$ & $64.0$ & $76.3$ & $84.3$     \\
            \specialrule{.08em}{0em}{0em}
            $\text{VGG}_{\text{F}}$                         & $96.3$ & $94.0$ & $37.6$ & $54.8$ & $35.6$ & $72.9$     \\
            \hline
            $\text{VGG}_{\text{MP}}$                        & $97.7$ & $96.0$ & $61.3$ & $68.3$ & $\bm{88.4}$ & $\bm{94.0}$     \\
            \specialrule{.08em}{0em}{0em}
            $\text{AlexNet} H_{0, 30}$                      & $79.8$ & $93.8$ & $58.0$ & $59.3$ & $69.8$ & $72.5$     \\
            \hline
            $\text{AlexNet} H_{1, 30}$                      & $96.2$ & $\bm{96.5}$ & $36.5$ & $54.3$ & $34.6$ & $65.4$     \\
            \hline
            $\text{AlexNet} H_{1, 30}^{\text{v}}$           & $98.0$ & $93.8$ & $61.7$ & $59.7$ & $66.1$ & $82.0$     \\
            \hline
            $\text{AlexNet} H_{2, 30}^{\text{diag}}$        & $96.7$ & $95.8$ & $41.0$ & $68.2$ & $34.6$ & $76.5$     \\
            \specialrule{.08em}{0em}{0em}
            $\text{VGG} H_{0, 30}$                          & $81.3$ & $90.8$ & $\bm{69.7}$ & $72.0$ & $74.6$ & $82.1$     \\
            \hline
            $\text{VGG} H_{1, 30}$                          & $92.5$ & $92.5$ & $60.2$ & $72.5$ & $33.6$ & $77.4$     \\
            \hline
            $\text{VGG} H_{1, 30}^{\text{v}}$               & $\bm{98.6}$ & $95.5$ & $66.3$ & $67.2$ & $62.9$ & $86.2$     \\
            \hline
            $\text{VGG} H_{2, 30}^{\text{diag}}$                          & $97.2$ & $96.3$ & $54.5$ & $\bm{74.8}$ & $47.6$ & $84.3$     \\
            \specialrule{.08em}{0em}{0em}
            $\text{AlexNet} H_{1, 20}^{\text{v, red}\mathcal{P}}$      & $95.7$ & $93.7$ & $38.8$ & $47.2$ & $56.2$ & $76.3$     \\
            \hline
            $\text{AlexNet} H_{2, 20}^{\text{v, red}\mathcal{P}}$      & $97.2$ & $95.0$ & $36.8$ & $50.7$ & $42.3$ & $67.8$     \\
            \hline
            $\text{VGG} H_{1, 20}^{\text{v, red}\mathcal{P}}$          & $95.2$ & $92.0$ & $38.5$ & $49.0$ & $46.1$ & $65.8$     \\
            \hline
            $\text{VGG} H_{2, 20}^{\text{v, red}\mathcal{P}}$          & $98.2$ & $96.5$ & $38.0$ & $56.2$ & $44.8$ & $74.0$     \\
            \specialrule{.08em}{0em}{0em}
            $\text{AlexNet} H_{1, 30}^{\text{v, red}F}$      & $98.0$ & $93.0$ & $40.0$ & $45.0$ & $33.1$ & $35.1$     \\
            \hline
            $\text{AlexNet} H_{2, 10}^{\text{v, red}F}$      & $98.4$ & $88.8$ & $37.2$ & $44.0$ & $33.1$ & $35.1$     \\
            \hline
            $\text{AlexNet} H_{2, 20}^{\text{v, red}F}$      & $98.1$ & $92.0$ & $38.3$ & $42.7$ & $33.4$ & $35.9$     \\
            \hline
            $\text{VGG} H_{1, 30}^{\text{v, red}F}$          & $\bm{98.6}$ & $95.7$ & $39.2$ & $40.7$ & $33.0$ & $35.0$     \\
            \hline
            $\text{VGG} H_{2, 10}^{\text{v, red}F}$          & $98.1$ & $94.8$ & $37.2$ & $40.8$ & $32.8$ & $35.5$     \\
            \hline
            $\text{VGG} H_{2, 20}^{\text{v, red}F}$          & $97.9$ & $94.5$ & $36.2$ & $41.7$ & $33.0$ & $35.6$     \\
            \hline
        \end{tabular}}
    \caption{Percentage accuracy for classification with CNN fingerprints as input, with highest accuracies emphasised in bold for each classification method.}
    \label{tab:cnn-results}
\end{table}

\end{document}